%% file: shared_interest.tex
\documentclass[manuscript,sigconf]{acmart}
\AtBeginDocument{%
  \providecommand\BibTeX{{%
    \normalfont B\kern-0.5em{\scshape i\kern-0.25em b}\kern-0.8em\TeX}}}

\copyrightyear{2022} 
\acmYear{2022} 
\setcopyright{rightsretained} 
\acmConference[CHI '22]{CHI Conference on Human Factors in Computing Systems}{April 29-May 5, 2022}{New Orleans, LA, USA}
\acmBooktitle{CHI Conference on Human Factors in Computing Systems (CHI '22), April 29-May 5, 2022, New Orleans, LA, USA}
\acmDOI{10.1145/3491102.3501965}
\acmISBN{978-1-4503-9157-3/22/04}

\usepackage{multirow}

\begin{document}
\title[Shared Interest]{Shared Interest: Measuring Human-AI Alignment to Identify Recurring Patterns in Model Behavior}

\author{Angie Boggust}
\email{aboggust@mit.edu}
\orcid{0000-0002-9671-5574}
\affiliation{%
  \institution{MIT CSAIL}
  \city{Cambridge}
  \state{Massachusetts}
  \country{USA}
}

\author{Benjamin Hoover}
\affiliation{%
  \institution{IBM Research}
  \city{Cambridge}
  \state{Massachusetts}
  \country{USA}
}

\author{Arvind Satyanarayan}
\affiliation{%
  \institution{MIT CSAIL}
  \city{Cambridge}
  \state{Massachusetts}
  \country{USA}
}

\author{Hendrik Strobelt}
\affiliation{%
  \institution{IBM Research}
  \city{Cambridge}
  \state{Massachusetts}
  \country{USA}
}

\renewcommand{\shortauthors}{Boggust et al.}

\begin{abstract}
    \input{sections/00_abstract}
\end{abstract}

\ccsdesc[500]{Computing methodologies~Machine learning}
\ccsdesc[500]{Human-centered computing~Human computer interaction (HCI)}

\keywords{human-computer interaction, interpretability, machine learning, saliency methods}

\maketitle
\input{sections/01_introduction} 
\input{sections/02_relatedwork}
\input{sections/03_method}
\input{sections/04_cases}
\input{sections/05_workflows}
\input{sections/06_discussion}

\begin{acks}
This work is supported by a grant from the MIT-IBM Watson AI Lab.
Research was also sponsored by the United States Air Force Research Laboratory and the United States Air Force Artificial Intelligence Accelerator and was accomplished under Cooperative Agreement Number FA8750-19-2-1000. The views and conclusions contained in this document are those of the authors and should not be interpreted as representing the official policies, either expressed or implied, of the United States Air Force or the U.S. Government. The U.S. Government is authorized to reproduce and distribute reprints for Government purposes notwithstanding any copyright notation herein.
\end{acks}

\bibliographystyle{ACM-Reference-Format}
\bibliography{shared_interest}

\appendix
\input{sections/07_appendix}

\end{document}

%% file: sections/00_abstract.tex
Saliency methods\,---\,techniques to identify the importance of input features on a model's output\,---\,are a common step in understanding neural network behavior.
However, interpreting saliency requires tedious manual inspection to identify and aggregate patterns in model behavior, resulting in ad hoc or cherry-picked analysis.
To address these concerns, we present Shared Interest: metrics for comparing model reasoning (via saliency) to human reasoning (via ground truth annotations).
By providing quantitative descriptors, Shared Interest enables ranking, sorting, and aggregating inputs, thereby facilitating large-scale systematic analysis of model behavior.
We use Shared Interest to identify eight recurring patterns in model behavior, such as cases where contextual features or a subset of ground truth features are most important to the model. 
Working with representative real-world users, we show how Shared Interest can be used to decide if a model is trustworthy, uncover issues missed in manual analyses, and enable interactive probing. 

%% file: sections/01_introduction.tex
\section{Introduction}
As machine learning continues to be deployed in real-world applications, it is increasingly important to understand the reasoning behind model decisions.
A common first step for doing so is to compute the model's \emph{saliency}.
In this setting, saliency is the output of any function that, given an input instance (e.g., an image), computes a score representing the importance of each input feature (e.g., pixel) to the model's output. 
Example saliency methods range from Vanilla Gradients~\citep{simonyan2013deep}, where scores represent the amount a small change in an input feature would have on the model's output, to black-box methods like LIME~\citep{ribeiro2016should} that use interpretable surrogate models trained to mimic the original model's decision boundary.
By analyzing saliencies, users can identify features important to the model's decision and determine how aligned these features are with human decision-making.

While saliency methods provide the much-needed ability to inspect model behavior, making sense of their output can still present analysts with a non-trivial burden.
In particular, saliencies are often visualized as solitary heatmaps, which do not provide any additional structure or higher-level visual abstractions to aid analysts in interpretation. 
As a result, analysts must rely solely on their visual perception and priors to generate hypotheses about model behavior. 
Similarly, saliency methods operate on individual instances, making it difficult to conduct large-scale analyses of model behavior and uncover recurring patterns. 
As a result, analysts must choose between time-consuming (often infeasible) manual analysis of all instances or ad hoc (often biased) selection of meaningful subsets of instances.

In response, we introduce Shared Interest: a method for comparing model saliencies with human-generated ground truth annotations.
Shared Interest quantifies the alignment between these two components by measuring three types of coverage: Ground Truth Coverage (GTC), or the proportion of ground truth features identified by the saliency method; Saliency Coverage (SC), or the proportion of saliency features that are also ground truth features; and IoU Coverage (IoU), the similarity between the saliency and ground truth feature sets.
These coverage metrics enable a richer and more structured interactive analysis process by allowing analysts to sort, rank, and aggregate input instances based on model behavior.
The metrics are agnostic to model architecture, input modality, and saliency method, and they can also be composed together (e.g., high SC and low GTC) to identify recurring patterns in the alignment of model and human decision-making. 

We demonstrate how Shared Interest enables structured large-scale analysis of model behavior across multiple domains and saliency methods. 
By applying Shared Interest to computer vision and natural language classification and regression tasks and using a variety of common saliency methods, we identify 8 recurring patterns of interesting model behaviors: \textsc{human aligned}, \textsc{sufficient subset}, \textsc{sufficient context}, \textsc{context dependent}, \textsc{confuser}, \textsc{insufficient subset}, \textsc{distractor}, and \textsc{context confusion}.
These patterns range from cases where the model's decision and explanation tightly align with human reasoning (\textsc{human-aligned}) to cases where contextual features are most important to the model's incorrect prediction (\textsc{distractor}).
Through representative case studies of real-world model interpretability workflows, we explore how Shared Interest helps a dermatologist and a machine learning researcher conduct more systematic analyses of model behavior. 
Users find that, unlike their prior exploration that was tedious and ad hoc, Shared Interest rapidly surfaces reasons to question a model's reliability, opportunities to learn from the model's representations, and issues missed during previous manual analysis.

We further demonstrate that Shared Interest is not only valuable to understanding a model's predictive performance but can also be used to \emph{query} model behavior. 
Leveraging the Shared Interest metrics alongside interactive human annotation enables a question-and-answer process where analysts probe input features and Shared Interest identifies the model's decisions whose saliency feature sets are most aligned. 
In an example human annotation workflow with an image classification task, we show how Shared Interest can reveal insights about the input features most salient to particular predictions and the model's understanding of secondary objects or background features.

Shared Interest is publicly available, with source code at \url{https://github.com/mitvis/shared-interest} and live demos at
\url{http://shared-interest.csail.mit.edu/}.

%% file: sections/02_relatedwork.tex
\section{Related Work}
Machine learning systems are increasingly designed for high-stakes tasks such as cancer diagnosis, and, as these systems achieve human-caliber or super-human accuracy~\citep{esteva2017dermatologist}, the temptation to deploy them correspondingly increases. 
In tandem, a body of work has identified dangerous pitfalls in commonly used models and their underlying training data~\citep{prabhu2020large, carter2020overinterpretation}.
To protect against the repercussions of deploying biased or ungeneralizable models, a growing effort focuses on understanding model decisions~\citep{doshivelez2017rigorous, rai2020explainable} and characterizing model errors~\citep{meek2016characterization}.
In this paper, we focus on post hoc saliency methods, also known as feature attribution methods~\citep{sturmfels2020visualizing}, that allow us to observe model reasoning~\citep{erhan2009visualizing, smilkov2017smoothgrad, springenberg2014striving, sundararajan2017axiomatic, lundberg2017unified, selvaraju2017grad, ribeiro2016should, carter2019made}.

Saliency methods explain deep learning model decisions on the instance-level.
Providing one interpretation at a time may be sufficient to answer questions about model behavior for a small collection of instances. However, it does not scale to answering questions about global model behavior or dataset characteristics.
Moreover, the output saliency maps require careful visual assessment to determine if the model used human-salient features to make its decision.
Together, these drawbacks often result in the tedious inspection of only a few examples that are cherry-picked or selected ad hoc.
By quantifying instances based on the agreement between model and human reasoning, Shared Interest offers a more comprehensive overview of model behavior across all instances and enables systematic evaluation of model behavior.

A recent body of work has questioned whether saliency methods are a reliable instrument for interpreting deep learning models~\citep{adebayo2018sanity, tomsett2020sanity, adebayo2020debugging, yang2019benchmarking, kindermans2019reliability}.
These papers propose saliency ``tests'' to measure each method's ability to faithfully represent model behavior.
While confirming the fidelity of saliency methods is a critical area of research, it is an orthogonal issue to the focus of our paper as even the most accurate saliency method will still exhibit instance-wise limitations.

Similar to our contribution are \citet{olah2018building} and \citet{kim2018interpretability} who argue feature-level saliency is not semantically meaningful enough and we should use higher levels of abstraction (e.g., hidden layer representations or concepts) instead. 
To combat the scalability limitations of instance-wise interpretation, they suggest decomposing activations through matrix factorization~\citep{olah2018building}, activation atlases~\citep{carter2019activation}, or concept activation vectors~\citep{kim2018interpretability}.
Shared Interest shares its underlying motivation with this work\,---\,a lack of semantically-meaningful structure in saliency methods and supporting scalable interpretability\,---\,but offers an alternate way forward.
In particular, although we compute attribution back to input features, we do so to compare salient features to human-provided ground truth. 
In doing so, Shared Interest brings structure and scale to the task of reading model saliencies and more directly expresses the alignment between human and model reasoning.

Aside from saliency methods, a growing number of techniques help users visually interpret models~\citep{hohman2018visual, wexler2019if}; however, these tools often focus on understanding patterns learned by intermediate nodes~\citep{bau2017network, zeiler2014visualizing, hohman2019summit} or are architecture-specific~\citep{kahng2018gan, hoover2020exbert, strobelt2016visual}.
In contrast, Shared Interest is agnostic to model architecture, saliency method, and dataset modality, and it can be incorporated into existing model interpretation workflows.

%% file: sections/03_method.tex
\section{The Shared Interest Method}
\label{sec:method}

Shared Interest is a method for computing the alignment between model and human decision-making. 
To do so, we introduce three metrics that measure the relationship between saliency and ground truth annotations.
We utilize these metrics to understand model behavior across computer vision (CV) and natural language processing (NLP) tasks.

\subsection{Metric Definitions}
\label{sec:metrics}

Mathematically, we use $S$ to represent the set of input features important for a model's decision as determined by a saliency method and $G$ to represent the set of input features important to a human's decision as indicated by a ground truth annotation.
For example, in a CV classification task, $G$ might represent the pixels within an object-level bounding box, and $S$ might represent the set of pixels salient to the model's decision as determined by a saliency method.
Similarly, in an NLP sentiment classification task, $G$ might be the set of input tokens annotated as indicative of sentiment, and $S$ is the set of tokens determined to be important to the model's prediction.

\input{figures/metrics}

We compute three metrics: IoU Coverage (IoU), Ground Truth Coverage (GTC), and Saliency Coverage (SC).
Each metric takes $G$ and $S$ as inputs and outputs a score between 0 and 1, inclusive.
\begin{align}
    \text{IoU} &= \frac{|G \cap S|}{|G \cup S|} 
    \label{eq:iou}
    \\
    \text{GTC} &= \frac{|G \cap S|}{|G|}
    \label{eq:gtc}
    \\
    \text{SC} &= \frac{|G \cap S|}{|S|}
    \label{eq:sc}
\end{align}
IoU (Eq.~\ref{eq:iou}) is the strictest metric, and it represents the similarity between the ground truth and saliency feature sets.
It is the number of features in both the ground truth and saliency sets divided by the number of features in at least one of the ground truth and saliency sets.
In machine learning terms, it is the Jaccard index.
GTC (Eq.~\ref{eq:gtc}) measures how strictly the model relies on \emph{all} ground truth features\,---\,the proportion of the ground truth feature set, $G$, that is also part of the saliency feature set, $S$.
It is analogous to concepts of recall or sensitivity in machine learning: the proportion of true positives (saliency features that are also ground truth features) successfully identified among all positives (ground truth features).
SC (Eq.~\ref{eq:sc}) measures how strictly the model relies on \emph{only} ground truth features\,---\,the proportion of the saliency feature set, $S$, that is also part of the ground truth feature set, $G$.
In machine learning terms, it is analogous to precision: the fraction of true positives (saliency features that are also ground truth features) successfully identified among all detected positives (saliency features).

\input{figures/metric_examples}

A score of zero under all three metrics means that an instance's saliency and ground truth feature sets are disjoint, which often indicates that background information is important to the model's prediction.
In Figure~\ref{fig:metric-grid}, we show example scenarios using an ImageNet~\citep{deng2009imagenet} image classification task and LIME~\citep{ribeiro2016should} saliency maps (see Section~\ref{sec:experiment-details} for details).
When a correctly classified instance has a low score, it often indicates there is contextual information in the background that is important to the model, such as the train tracks surrounding the \emph{electric locomotive}. 
When an instance has a low score and is incorrectly classified, it can indicate the model is focusing on a secondary object (e.g., the wrong dog) or incorrectly relying on background context (e.g., using snow to predict \emph{arctic fox}).

A high IoU score indicates the explanation and ground truth feature sets are very similar ($\text{IoU}=1 \implies S = G$), meaning the features that are critical to human reasoning are also important to the model's decision.
Correctly classified instances with high IoU scores indicate the model was correct in ways that tightly align with human reasoning. 
Incorrectly classified instances with high IoU scores, on the other hand, are often challenging for the model, such as the image of a snowplowing truck that is labeled as \emph{snowplow} but predicted as \emph{pickup}.

High GTC signals that the ground truth features are the most relevant to the model's decision ($\text{GTC}=1 \implies G \subseteq S$).
When a correctly classified instance has high GTC it indicates that the model relies on the object and relevant background features (e.g., the cab and the street) to make a correct prediction.
Incorrectly classified instances with high GTC are examples where the model overly relies on local contextual information such as using the keyboard and person's lap to predict \emph{laptop}.

High SC indicates the model relies almost exclusively on ground truth features to make its prediction ($\text{SC}=1 \implies S \subseteq G$).
Filtering for correctly classified instances with high SC can surface instances where a subset of the object, such as the dog's face, was important to the model's prediction.
Incorrectly classified instances with high SC suggests that an insufficient portion of the object is salient to the prediction (e.g., a small region of black and white spots to predict \emph{dalmatian}).

Shared Interest metrics can also be combined to yield exciting insights.
For example, instances with high SC and low GTC indicate the model is focused on a subset of the ground truth region, whereas high GTC and low SC indicate the model is relying on the ground truth and contextual features to make its prediction.

\subsection{Experimental Setup}
\label{sec:experiment-details}

In subsequent sections of this paper, we apply Shared Interest to CV and NLP tasks, including multi-class image classification, binary classification of medical images, and sentiment regression on text reviews.
Shared Interest surfaces interesting results across a variety of saliency methods, including gradient-based methods like Vanilla Gradients~\citep{simonyan2013deep} and Integrated Gradients~{\citep{sundararajan2017axiomatic}}, as well as model-agnostic methods like LIME~{\citep{ribeiro2016should}} and SIS~{\citep{carter2019made}}.
We explore additional saliency methods in Section~\ref{sec:appendix-saliency}.
Since $S$ is a discrete feature set, Shared Interest can be straightforwardly applied to saliency methods like SIS~{\citep{carter2019made}} that output feature sets.
However, to apply Shared Interest to methods that output a continuous score (e.g., Integrated Gradients~{\citep{sundararajan2017axiomatic}}), we compute $S$ by discretizing these scores. 
We demonstrate that Shared Interest is robust to discretization procedure by employing score-based and model-based thresholding.
Score-based thresholding, used in the CV examples, creates discrete feature sets using only the saliency.
For example, we threshold Vanilla Gradients at one standard deviation above the mean to allow for variance in the number and value of salient features across instances, and we select LIME's top $n$ positively contributing features to demonstrate that even naive thresholding can be effective.
Model-based thresholding, used in the NLP examples, creates discrete feature sets containing features directly correlated with the model's prediction.
In these examples, features positively correlated with the model's prediction are iteratively selected until the model can confidently predict the correct class using only those features.
Section~\ref{sec:appendix-thresholding} contains additional examples of discretization techniques.

\paragraph{ImageNet Image Classification}
In the ImageNet image classification examples (Sections~{\ref{sec:metrics}},~{\ref{sec:cases}}, and~{\ref{sec:what-if}}), we use two subsets of the original ImageNet dataset: the dog and vehicle subsets from ImageNet-9~{\citep{xiao2020noise}}.
Since ImageNet only provides bounding box annotations for a subset of images, we further subset these sets to only contain images with annotations.
We use features in the bounding box regions as $G$.
We use a pretrained ResNet50~{\citep{he2016deep}} provided by PyTorch~{\citep{paszke2019pytorch}} trained on 1000-way classification on ImageNet.
In Sections~{\ref{sec:metrics}}~and~{\ref{sec:cases}}, we use LIME~{\citep{ribeiro2016should}} explanations as $S$.
To compute LIME, we use the author's implementation (\url{https://github.com/marcotcr/lime}) with 1000 samples per image, a Ridge Regression linear model, cosine distance function, and an exponential kernel.
We create the saliency feature set using the top 5 features that had a positive impact on the model's prediction, where features are super-pixels defined by QuickShift~{\citep{vedaldi2008quick}}.
In Section~{\ref{sec:what-if}}, we compute Vanilla Gradients~\citep{erhan2009visualizing} using Captum~\citep{kokhlikyan2020captum} for all 1000 ImageNet classes.
We take the absolute value of the gradients and discretize by thresholding each saliency map at one standard deviation above the mean.

\paragraph{Melanoma Classification}
In the melanoma classification example (Section~{\ref{sec:dermatologist}}), we use lesion images and segmentations from the ISIC 2016 Challenge~{\citep{gutman2016skin}}.
Each image is classified as malignant or benign and contains a lesion segmentation that we use to represent $G$.
We trained a ResNet50~{\citep{he2016deep}} model from scratch for 4 epochs using Cross-Entropy loss, Adam~{\citep{kingma2014adam}} optimization, a learning rate of 0.1, a batch size of 128, and class-weighted sampling.
Since the test set is not public, we evaluate on the validation set and achieve 0.822 balanced class accuracy.
We use LIME~{\citep{ribeiro2016should}} explanations as $S$.
To compute LIME, we use the author's implementation (\url{https://github.com/marcotcr/lime}) with 1000 samples per image, a Ridge Regression linear model, cosine distance function, and an exponential kernel.
We create the saliency feature set using the top 5 features that had a positive impact on the model's prediction, where features are super-pixels defined by QuickShift~{\citep{vedaldi2008quick}}.

\paragraph{BeerAdvocate Sentiment Regression}
In the beer review regression examples (Sections~{\ref{sec:cases}} and~{\ref{sec:ml-researcher}}), we use beer reviews from the BeerAdvocate dataset processed by \citet{lei2016rationalizing}.
Each review is annotated with scores ranging from 0 to 1 in 0.1 increments representing 0 to 5 star reviews in half-star increments.
Each review has a score and sentence level annotation for each aspect (aroma, appearance, palette, taste).
To directly compare Shared Interest to prior saliency method analysis (Section~{\ref{sec:ml-researcher}}), we use the recurrent neural networks (RNNs), SIS rationales, and integrated gradient explanations from \citet{carter2019made} available at: \url{https://github.com/b-carter/SufficientInputSubsets}.
The RNNs were trained on each aspect of the dataset.
The SIS procedure selected the sufficient input subsets using an 85\% model confidence threshold.
For direct comparison, the Integrated Gradients were also iteratively selected from highest to lowest impact on the predicted class until the model made the original prediction with 85\% confidence.
To apply the Shared Interest definitions to this regression task, we define correctness as whether the model's output is within a half-star ($\pm \Delta = 0.05$) of the actual value.

%% file: figures/metrics.tex
\begin{figure}[t]
      \centering
      \centerline{\includegraphics[width=\linewidth]{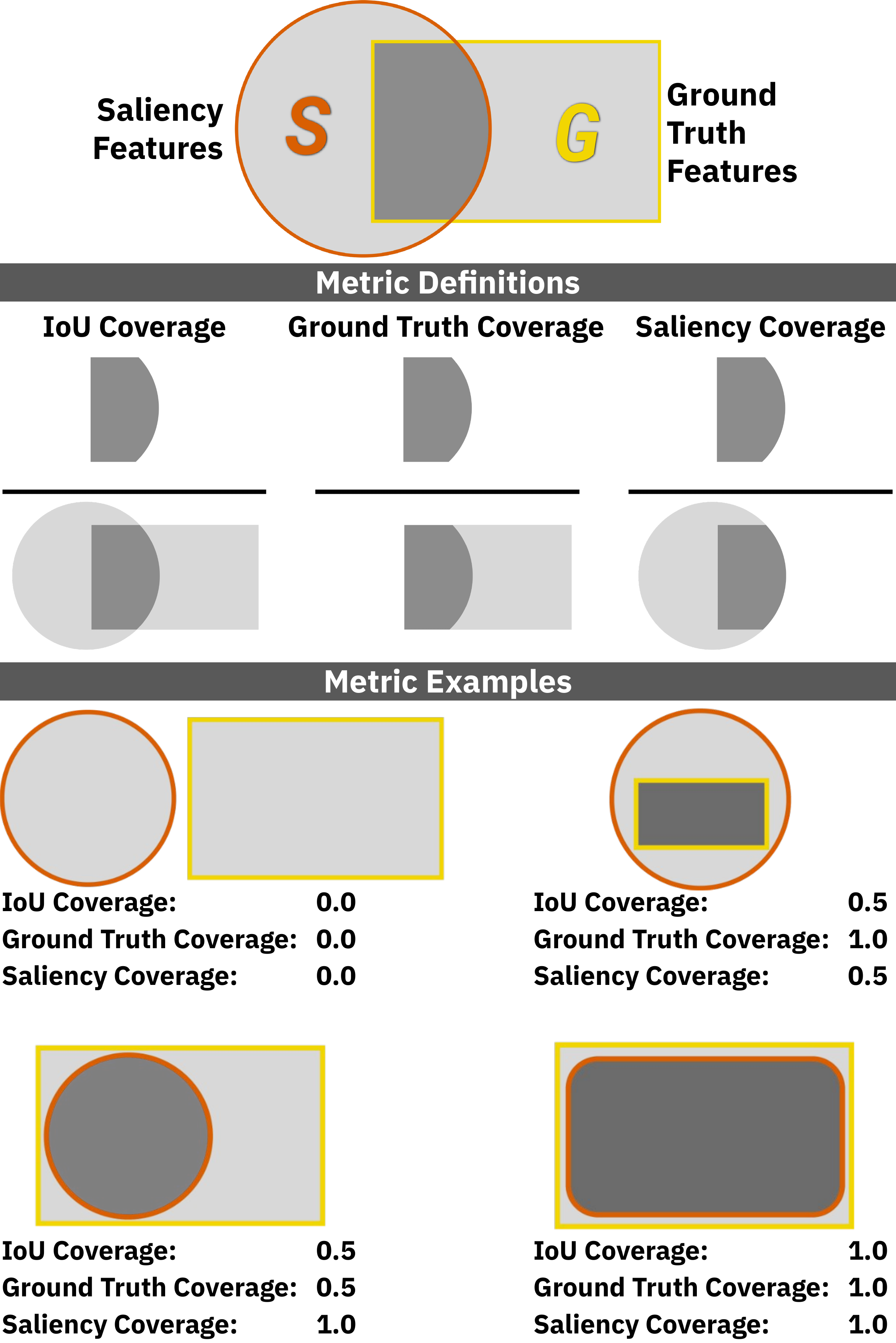}}
      \caption{Shared Interest takes a set of saliency features $S$ and a set of ground truth features $G$ and outputs three metrics for identifying instances of interest: IoU, GTC, SC. IoU represents the alignment of model-salient and human-salient features. GTC represents the proportion of human-salient features used by the model. SC represents the proportion of model-salient features used by a human.}
      \label{fig:metrics}
      \Description{Schematic of the Shared Interest metrics. Saliency features are displayed as an orange shape, and ground truth features are shown as a yellow shape. The shapes overlap. Each metric is shown as a fraction. IoU Coverage is shown as the intersection of the shapes over the union of the shapes. Ground Truth Coverage is shown as the intersection of the shapes over the yellow shape. Saliency Coverage is shown as the intersection of the shapes over the orange shape. Four examples are shown. The first shows disjoint orange and yellow shapes. The text below shows IoU Coverage: 0.0, Ground Truth Coverage 0.0, Saliency Coverage: 0.0. The second example shows a yellow shape contained within an orange shape. The text below shows IoU Coverage: 0.5, Ground Truth Coverage 1.0, Saliency Coverage: 0.5. The third example shows an orange shape contained within a yellow shape. The text below shows IoU Coverage: 0.5, Ground Truth Coverage 0.5, Saliency Coverage: 1.0. The fourth example shows completely overlapping orange and yellow shapes. The text below shows IoU Coverage: 1.0, Ground Truth Coverage 1.0, Saliency Coverage: 1.0.}
\end{figure}

%% file: figures/metric_examples.tex
\begin{figure}[t]
      \centering
       \centerline{\includegraphics[width=\linewidth]{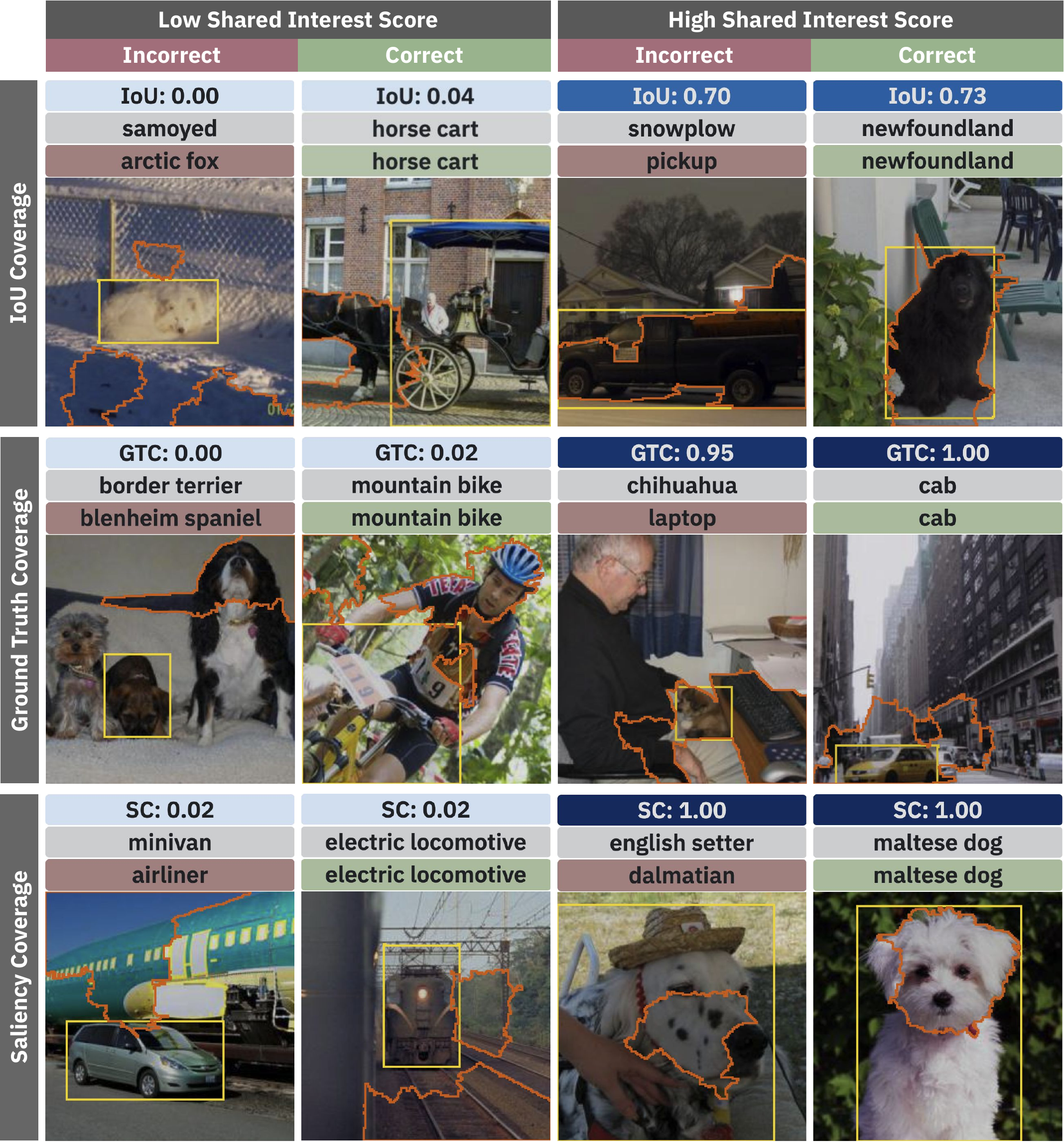}}
      \caption{The Shared Interest metrics uncover interesting instances of model behavior. Here, we show an example ImageNet images with high and low scores for each Shared Interest metric. Each image is annotated with its label (grey), prediction (green if correct, red otherwise), ground truth features (yellow), and LIME saliency features (orange). A score of zero under all three metrics indicates the ground truth set ($G$) and saliency feature set ($S$) are disjoint. High scores can indicate the model is relying on the ground truth features (IoU), a subset of the ground truth features (SC), or a superset of the ground truth features (GTC). }
      \label{fig:metric-grid}
      \Description{A grid of image examples showing correct and incorrectly classified images with high and low scores for each Shared Interest metric. The rows are IoU Coverage, Ground Truth Coverage, and Saliency Coverage. The columns are low score and incorrect, low score and correct, high score and incorrect, low score and correct. The low IoU Coverage and incorrect cell shows an image of a Samoyed classified as an arctic fox. The saliency and ground truth regions are disjoint. Low IoU Coverage and correct shows a correctly classified image of a horsecart. The ground truth features contain the cart, and the saliency features contain the horse. High IoU Coverage and incorrect shows an image of a snowplow classified as a pickup. The ground truth and saliency features both contain the vehicle. High IoU Coverage and correct shows a correctly classified image of a Newfoundland dog. The ground truth and saliency features both contain the dog. Low Ground Truth Coverage and incorrect shows an image of three dogs labeled as Border Terrier and predicted as Blenheim Spaniel. The ground truth and saliency regions indicate different dogs. Low Ground Truth Coverage and correct shows a correctly classified mountain bike image. The ground truth features contain the bike, and the saliency features contain the bicyclist. High Ground Truth Coverage and incorrect shows an image of a man at a desk with a dog in his lap. The image is labeled as Chihuahua and classified as a laptop. The ground truth features focus on the dog, and the saliency features focus on the dog, desk, and keyboard. High Ground Truth Coverage and correct shows a correctly classified image of a cab. The ground truth features focus on the cab, and the saliency features focus on the cab and surrounding street. Low Saliency Truth Coverage and incorrect shows an image of a van in front of a large vehicle. The image is labeled as a minivan and classified as an airliner. The ground truth features contain the van, and the saliency features focus on the background vehicle. Low Saliency Coverage and correct shows a correctly classified image of an electric locomotive. The ground truth features contain the locomotive, and the saliency features contain the locomotive tack. High Saliency Coverage and incorrect shows an image of a dog labeled as English Setter and classified as Dalmatian. The ground truth features focus on the dog, and the saliency features focus on the spots on the dog's face. High Saliency Coverage and correct shows a correctly classified image of a Maltese dog. The ground truth features focus on the entire dog, and the saliency features focus only on the dog's head.}
\end{figure}

%% file: sections/04_cases.tex
\section{Recurring Patterns in Model Behavior}
\label{sec:cases}

To study how Shared Interest could aid in understanding model behavior, we conducted iterative rounds of qualitative analysis. 
We applied our metrics to models across a range of domains (computer vision and natural language processing), tasks (regression or classification), saliency methods (gradient-based or model-agnostic), and model architectures (convolutional or recurrent neural networks).
Using Shared Interest to sort, explore, and characterize individual instances, several patterns in model decision-making emerged.
We validated these patterns and refined their definitions through further iterative analysis (i.e., applying Shared Interest to additional datasets and models and confirming that these patterns continued to surface). 
Ultimately, we identified eight cases of model behavior defined in terms of Shared Interest metrics and model correctness. 
In Figure~{\ref{fig:cases}}, we show an example of each case in a computer vision setting (ImageNet classification with LIME) and a natural language processing setting (BeerAdvocate aroma sentiment prediction~\citep{mcauley2012learning,lei2016rationalizing, carter2019made} with Integrated Gradients~\citep{sundararajan2017axiomatic, carter2019made}).

\input{figures/cases}

\subsection{Human Aligned}
Instances that fall into the \textsc{human aligned} category are predicted correctly and have high IoU, thus indicating that the model is making a correct prediction and its rationale for that prediction aligns with a human's.
For example, in the CV setting of Figure~\ref{fig:cases}a, almost every pixel in the ground truth bounding box was important for the model to make the correct prediction of \emph{trailer truck}.
In the NLP example, the saliency method shows that the model uses almost every word related to the beer's aroma to make its correct prediction of 0.9 (strong positive sentiment).
\textsc{human aligned} instances indicate cases when the model is faithful to human decisions, and ideally, all instances would fall into this category.

\subsection{Sufficient Subset}
The \textsc{sufficient subset} category contains instances with high SC and low GTC, revealing where a subset of the human-annotated features are important for the model to make a correct prediction.
For example, in Fig.~\ref{fig:cases}(b), the human-annotated ground truth in the CV example covers the entire tractor. 
However, the saliency method indicates that the tractor's tire was most important for the model to make its correct prediction.
In the NLP case, the salient regions indicate the model considers the words ``complex'', ``aroma'', ``chocolate'', and ``vanilla'' important to predict strong positive sentiment.
The \textsc{sufficient subset} category may indicate that the human reasoning annotation includes extraneous information (e.g., the stop words in the aroma review) or that the model relies on an adequate but incomplete set of features that may not generalize (e.g., only the tractor tire).

\subsection{Sufficient Context}
The \textsc{sufficient context} category includes correctly predicted instances with low IoU, indicating there is information in non-ground truth features correlated with the correct prediction.
Analyzing instances in this category can validate if contextual features are indeed meaningful and not spurious correlation.
The CV example in Figure~\ref{fig:cases}c shows the model uses the helmet to predict \emph{snowmobile}.
While the presence of a snowmobile helmet correlates with the existence of a snowmobile, this model would not be robust to real-life scenarios.
This result might inspire further exploration of instances with snowmobiles and snowmobile helmets to confirm there is not a rigid dependence between the two objects.
In the NLP example, the saliency method indicates that the word ``zilch'' is important to the model's decision.
However, ``zilch'' corresponds to the taste aspect instead of aroma.
This discrepancy suggests a correlation between reviewer scores for aroma and other aspects and that the model may overfit to features related to other aspects.
Instances in these cases may be of interest to an analyst to identify correlated features that exist in the training data but would not generalize to the real world.

\subsection{Context Dependent}
The \textsc{context dependent} category identifies correctly classified instances with high GTC and low SC, meaning the model relies on ground truth and contextual features to make a correct prediction.
In Figure~\ref{fig:cases}d, the CV example shows the model relies not only on the streetcar (the labeled class) but also on the train tracks to predict \emph{streetcar}.
In the NLP example, the model uses the ground truth words ``rich'' and ``smells'' along with words like ``nutty'', ``cocoa'', and ``almond'' to make its positive sentiment prediction.
While the context is semantically correlated with the ground truth in both examples, these instances may indicate nongeneralizable correlations and require further exploration to uncover whether it is reasonable for the model to use context in its prediction.

\subsection{Confuser}
Confusers are instances where the model relies on human-salient features but still makes an incorrect prediction.
In Shared Interest terms, members of the \textsc{confuser} case are incorrectly classified instances with high IoU.
In the CV example in Figure~\ref{fig:cases}e, the \textsc{confuser} case identifies an ambiguous label\,---\,the image is labeled as \emph{moped}, but the model predicts \emph{motor scooter}\,---\, a known problem with ImageNet~\citep{beyer2020we, tsipras2020imagenet}.
Using Shared Interest, instances of this failure case immediately rise to the forefront of the analysis process via the \textsc{confuser} class.
In the NLP example, the saliency method finds the model relies on almost all the ground truth words to make a strong positive sentiment prediction.
While the ground truth sentence has a positive sentiment, the reviewer only gave the aroma a 0.6 (weakly positive).
In both domains, the \textsc{confuser} case helps immediately identify instances with imprecise dataset labels.
Discovering such instances might encourage an analyst to conduct further exploratory analysis on the dataset or perform additional preprocessing to resolve ambiguities.

\subsection{Insufficient Subset}
\textsc{Insufficient subset} identifies incorrectly classified instances with high SC and low GTC, meaning a subset of the ground truth features is important to the model, but it makes an incorrect prediction.
In Fig.~\ref{fig:cases}(f), with the CV example, the model predicts the dog breed \emph{whippet} on an image of two whippets in a shopping cart.
The image is labeled as \emph{shopping cart}, and the saliency method indicates the model relied on the faces of the dogs as opposed to the pixels of the shopping cart.
In the NLP example, the saliency method indicates the model relies upon the words ``vague'' and ``odor'' in the aroma sentence to predict negative sentiment (0.3).
However, other words in the sentence not indicated as salient to the model do contain positive sentiment (e.g., ``caramel malt'' and ``noble hops'') and likely contributed to the actual label of 0.6.
In general, \textsc{insufficient subset} cases can signal to an analyst that the model is overly reliant on a small set of features and warrants further exploration.

\subsection{Distractor}
\textsc{Distractors} are cases where the model does not rely on ground truth features (low IoU) and makes an incorrect prediction.
In Fig.~\ref{fig:cases}(g), the CV example shows an instance labeled \emph{moped}, but the model predicts \emph{church} as the saliency covers pixels related to the church in the background.
In this case, the image contains multiple objects but only has a single label, which is a known flaw of ImageNet~\citep{beyer2020we, tsipras2020imagenet}.
In the NLP example, the saliency method indicates the model relies on the words ``metal'', ``corn'', and ``nothing awesome'' to predict negative sentiment.
While the known aroma words (``the smell is sweet malty lagery'') have positive sentiment, the model is distracted by negative sentiment words elsewhere in the review.
\textsc{Distractor} instances may indicate that the model is overfitting to the overall sentiment of the review rather than the specific sentiment associated with the aroma.

\subsection{Context Confusion}
The \textsc{context confusion} case contains instances where the model is using ground truth features but is confused by other features and, thus, makes an incorrect prediction.
In Shared Interest terms, these instances have high GTC and low SC.
For example, in the CV setting in Figure~\ref{fig:cases}h, the saliency indicates the presence of the field next to the trailer truck is important for the model to predict \emph{harvester}.
In the NLP example, the model relies on words in the aroma sentence as well those surrounding the sentence to predict strong positive sentiment (0.9) as opposed to weakly positive sentiment (0.6).
In this instance, many of the surrounding words contain positive sentiment (e.g., ``impressive'' and ``extremely''), which may have caused the model to predict more positively than the actual class.

%% file: figures/cases.tex
\begin{figure*}[t]
    \begin{center}
    \centerline{\includegraphics[width=\linewidth]{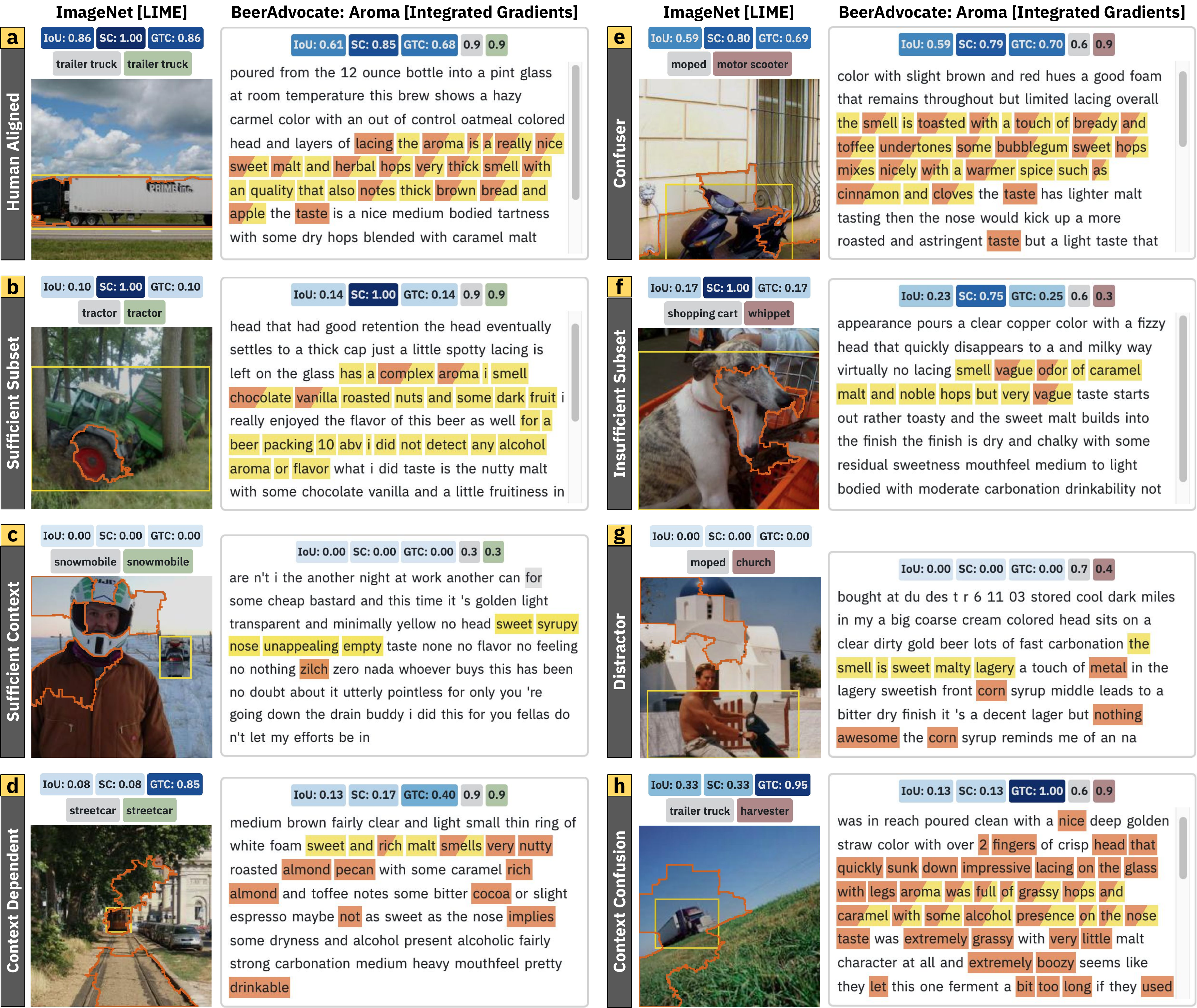}}
    \caption{Using Shared Interest we identify eight recurring patterns in model behavior across two domains and saliency methods: an ImageNet~\citep{deng2009imagenet} classification model with LIME~\cite{ribeiro2016should} and a BeerAdvocate sentiment regression model~\cite{carter2019made, mcauley2012learning,lei2016rationalizing} with Integrated Gradients~\cite{sundararajan2017axiomatic, carter2019made}. Each example includes the IoU, GTC, and SC scores, true label (grey), and model prediction (green if correct, red otherwise). Ground truth features are highlighted in yellow and saliency features are highlighted in orange.}
    \label{fig:cases}
    \Description{Each of the eight patterns of model behavior is labeled a-h. Each pattern is shown via a computer vision (CV) ImageNet example with LIME explanations and a natural language processing (NLP) BeerAdvocate Aroma example with integrated gradients. Case (a) shows human aligned examples. The CV example is an image of a correctly classified trailer truck where the saliency features highlight the truck. The NLP example is a correctly predicted positive beer review where the ground truth and saliency regions contain almost identical feature sets. Case (b) shows sufficient subset examples. The CV example is a correctly classified tractor image where the saliency feature set contains the tractor tire. The NLP example is a correctly predicted positive review where 4 of the 29 ground truth words are in the saliency feature set. Case (c) shows sufficient context examples. The CV example shows a correctly classified snowmobile image where the saliency feature set contains the snowmobiler's helmet. The NLP example is a correctly predicted negative beer review where the word ``zilch'' from elsewhere in the review is the saliency feature set. Case (d) shows context dependent examples. The CV example is a correctly classified streetcar image where the streetcar and tracks are in the saliency feature set. The NLP example shows a correctly predicted positive review where words in and near the ground truth sentence are in the saliency feature set. Case (e) shows confuser examples. The CV example is an image of a moped predicted to be a motor scooter. Its ground truth and saliency feature sets both cover the vehicle. The NLP example is a slightly positive review (0.6) predicted to be highly positive (0.9). Its saliency and ground truth feature sets contain almost all the same words. Case (f) shows insufficient subset examples. The CV example shows an image of dogs in a shopping cart classified as a shopping cart but predicted to be a whippet. The saliency features contain the dogs' faces. The NLP example is a slightly positive (0.6) review predicted to be negative (0.3). The saliency features contain the words ``vague'' and ``odor'' out of the 12-word sentence. Case (g) shows distractor examples. The CV example shows a moped in front of a church. It is classified as a moped but predicted to be a church. The saliency feature set contains the dome of the church. The NLP example is a positive review (0.7) predicted to be negative (0.4). Its saliency feature set and ground truth set are disjoint. Case (h) shows context confusion examples. The CV example shows an image of a trailer truck classified as a harvester. The saliency features contain the truck and the surrounding field. The NLP example is a slightly positive review (0.6) predicted to be highly positive (0.9). The saliency feature set contains the ground truth sentence and surrounding words.}
    \end{center}
\end{figure*}

%% file: sections/05_workflows.tex
\section{Interactive Interpretability Workflows}
We demonstrate how Shared Interest can be used for real-world analysis through case studies of three interactive interpretability workflows of deep learning models.
The first case study follows a domain expert (a dermatologist)
using Shared Interest to determine the trustworthiness of a melanoma prediction model.
The second case study follows a machine learning expert analyzing the faithfulness of their model and saliency method.
The final case study examines how Shared Interest can analyze model behavior even without pre-existing ground truth annotations.

We developed visual prototypes for each case study to make the Shared Interest method explorable and accessible to all users, regardless of machine learning background.
The computer vision and natural language processing prototypes (Figure~{\ref{fig:cv-prototype}} and Figure~{\ref{fig:nlp-prototype}}) focus on sorting and ranking input instances so users can examine model behavior.
Each input instance (image or review) is annotated with its ground truth features (highlighted in yellow) and its saliency features (highlighted in orange) and is shown alongside its Shared Interest scores, label, and prediction.
The interface enables sorting and filtering based on Shared Interest score, Shared Interest case, label, and prediction.
The human annotation interface (Figure~{\ref{fig:human-annotation-prototype}}) is designed for interactive probing.
The interface enables users to select and annotate an image with a ground truth region and returns the top classes with the highest Shared Interest scores for that ground truth region.
Code for the prototypes is available at \url{https://github.com/mitvis/shared-interest}, and live demos of each prototype are available at \url{http://shared-interest.csail.mit.edu/}.

\subsection{Model Analysis by a Domain Expert}
\label{sec:dermatologist}

Our first case study follows the use case of a domain expert, a dermatologist, who wishes to evaluate the trustworthiness of a machine learning model that could assist them in diagnosing melanoma.
Accurate and early melanoma diagnosis is a critical task that can significantly impact patient outcomes, and machine learning could assist dermatologists in making more accurate decisions.
In order to do so, however, our participant noted it would be imperative for dermatologists to be able to evaluate how the model operates personally.

\input{figures/cv_prototype}

We evaluate Shared Interest in this context to understand how its ability to convey model behavior may help a domain expert determine whether or not they should trust a model.
To do so, we applied Shared Interest to a Melanoma Classification task (see Section~\ref{sec:experiment-details} for details).
We used a Convolutional Neural Network trained on the ISIC Melanoma dataset~\citep{codella2018skin, tschandl2018ham} to classify images of lesions as either \emph{malignant} (cancerous) or \emph{benign}.
We used lesion segmentations from the dataset as the ground truth feature sets and the output of LIME~\citep{ribeiro2016should} towards the predicted class as the saliency feature sets.
Using a prototype visual interface (Figure~{\ref{fig:cv-prototype}}) designed to enable interactive analysis of Shared Interest, we explored examples with the dermatologist for 30 minutes.
We used the Shared Interest cases to outline the conversation by showing the dermatologist examples from each case. 
However, the dermatologist guided the analysis by discussing the insights that excited them and suggesting what to investigate next.
Throughout the conversation, we asked the dermatologist open-ended questions (e.g., \emph{``How do you feel about the model after seeing these examples?''}) to understand how they would evaluate a model and how Shared Interest could aid in evaluation.

Using the \textsc{human aligned}, \textsc{context dependent}, and \textsc{sufficient subset} categories, the dermatologist surfaced insight into cases where the model was trustworthy.
Analyzing \emph{malignant} lesions in the \textsc{human aligned} case surfaced examples where the model correctly classified cancerous lesions by relying on features of the lesion.
The dermatologist agreed with the model on these images and began to build trust with the model, noting \emph{``obviously it does a pretty good job.''}
\textsc{Context dependent} images identified cases where the model relied not only on the lesion but also on surrounding skin.
While there was potential for the dermatologist to distrust the model, they actually found these instances especially interesting because cancerous cells can lie beyond the pigmented lesion boundary. 
Thus, the dermatologist wondered if \emph{``there are really subtle changes that we are not picking up that [the model] is able to.''}
Images in the \textsc{sufficient subset} case showed cases where the model only relied on a subset of the lesion.
While the dermatologist agreed with the model, they expressed some concern that it was not using the complete lesion, especially when there were meaningful cancerous features in the unused regions. 

Shared Interest was also able to quickly reveal cases where the model was not trustworthy.
The \textsc{sufficient context} and \textsc{distractor} cases showed images where the model relied on contextual features such as peripheral skin regions or the presence of artifacts (see Figure~\ref{fig:cv-prototype}).
While the dermatologist was tolerant to a few instances where the model relied on non-salient features, seeing the number of images in these cases led the dermatologist to distrust the model in all cases, stating \emph{``I would discard the model.''}

By classifying inputs into cases where the model was or, more importantly, was not aligned with human reasoning, Shared Interest enabled the dermatologist to rapidly and confidently decide whether or not to trust the model. 
If the dermatologist had evaluated the model by randomly selecting images, they might not have identified that the model repeatedly made decisions based on background information, and they would not have known how frequently that case occurred.
As the dermatologist said, Shared Interest is \emph{``helpful [as a way to] see how the computer is thinking and allow me to understand if I should trust it.''}

\subsection{Saliency Method Analysis by a Machine Learning Researcher}
\label{sec:ml-researcher}
Our second case study is representative of use cases where a machine learning expert wants to analyze a model or saliency method they are developing.
To evaluate Shared Interest's value in the development pipeline, we worked with an author of the Sufficient Input Subset (SIS) interpretability method~\citep{carter2019made} whose goal is to understand how well SIS explains model decisions.
During development of the SIS method, one of the ways the researchers analyzed the method was by applying it to the BeerAdvocate dataset and comparing the SIS saliencies (called ``rationales'' by the researchers) to the ground truth annotations.
This process enabled them to evaluate whether the rationales \emph{``fell within the ground truth''} and represented a \emph{``compact set''} of meaningful features.

\input{figures/nlp_prototype}

To recreate the researcher's original workflow, we applied Shared Interest to the BeerAdvocate reviews annotated on the appearance aspect, trained Recurrent Neural Networks, and SIS rationales from \citet{carter2019made} (see Section~\ref{sec:experiment-details} for details).
We populated the visual prototype with the results (Figure~{\ref{fig:nlp-prototype}}) and used it to explore the Shared Interest cases with the researcher.
Throughout the 45 minute conversation, we asked the researcher open-ended questions to understand how they evaluate a saliency method and how Shared Interest might aid in their evaluation.
While we used the Shared Interest cases as a guide, the researcher led the analysis, and insights from the cases often inspired them to examine additional settings.

Using Shared Interest, the researcher surfaced numerous insights that inspired confidence in the SIS algorithm.
For example, the researcher immediately identified that most reviews have high SC, indicating most of the SIS rationales were contained almost entirely within the ground truth.
Since \emph{``ideally, the model is learning the right set of features and thus the rationales live within the correct set of features''}, the researcher found the distribution of scores indicative that the SIS procedure was capturing meaningful information.
In their original analysis, the researchers had even computed a metric equivalent to SC as a quantitative way to analyze their method. 
So, seeing the same metric populated by Shared Interest validated the use of Shared Interest and the \textsc{sufficient subset} and \textsc{insufficient subset} categories.
The researcher found it reassuring to find \textsc{human aligned} and \textsc{sufficient subset} instances that matched their expectations, such as rationales that contained appearance-specific words (e.g., ``red'', ``copper'', and ``head'') but did not contain uninformative ground truth words like stop words.
The \textsc{sufficient subset} category was significant to the researcher since it aligned with SIS's goal to find minimal rationales.
Seeing all of these examples at once helped the researcher identify cases where the rationale was indeed a meaningful sufficient subset of words such as ``lovely looking''.

Shared Interest also helped the researcher uncover previously unknown pitfalls in the model and the data.
Looking at instances in the \textsc{sufficient subset} category, the researcher identified common cases of model overfitting, such as a correct prediction using only the word ``beautiful''.
As the researcher put it, \emph{``These are positive words, so it makes sense they are correlated with positive appearance, but I don't think they should be sufficient for separating [the appearance] aspect from others.''}
Looking at reviews in the \textsc{insufficient context} case exposed instances where the model was again overfitting to positive sentiment. However, now the researcher was even more concerned since it caused an incorrect prediction.
Although the researcher had \emph{``previously observed that the model had associated single tokens that were general positive sentiment words with predicting high sentiment''}, they did not \emph{``as quickly notice particular words like `beautiful' that were immediately surfaced [via Shared Interest].''}
Finally, looking at \textsc{sufficient context} reviews, where SIS rationales are disjoint from the ground truth annotation, the researcher uncovered reviews with incomplete or incorrect annotations.
Until using Shared Interest, the researcher had previously never identified an incorrectly annotated review, saying \emph{``In the past, I did not note any cases where I thought the annotations might have been incomplete. I think that's a pretty interesting insight.''}

Overall, the researcher found that grouping and aggregating via Shared Interest helped them \emph{``see all of the [reviews] grouped together by the various cases''} which categorized and \emph{``clearly described what the various patterns are''}.
In the researchers' original analysis, they had \emph{``skimmed through a big file of [reviews] not sorted in any way''}, and, while they \emph{``were noting patterns, it was harder to keep track of these different cases.''}
If they would have had access to Shared Interest at the time of their original analysis, this researcher thought it would have \emph{``more quickly exposed some of the patterns and behaviors that we identified and also led to additional discoveries.''}

\subsection{Interactive Probing of Model Behavior}
\label{sec:what-if}

For our final case study, we demonstrate a workflow where Shared Interest can be used as a mechanism to \emph{query} model behavior.
For any input instance, rather than computing the saliency for only the predicted class, we do so for all possible classes.
Moreover, users can interactively annotate the instance to designate a ``ground truth'' region of interest instead of relying on a single pre-existing ground truth.
By calculating all three Shared Interest metrics between these two sets of features and returning classes with the highest Shared Interest scores, we enable users to engage in a style of ``what if'' reasoning.
Users can interactively probe the model to understand what input features are important to trigger a particular prediction.

\input{figures/human_annotation_prototype}

In Figure~\ref{fig:human-annotation-example}, we show an example of this style of ``what if'' analysis on an ImageNet classification task (see Section~\ref{sec:experiment-details}) for details).
By interactively re-specifying the ``ground truth'' on a single image, we repeatedly probe the model and surface insights about its behavior.
Since the model was trained to predict the \emph{otterhound} in the image, we can use Shared Interest to validate that the model has indeed learned the salient features of the dog. 
By selecting the pixels associated with the dog's face and body (Figure~\ref{fig:human-annotation-example}a), we find that, although none of the top three returned classes are \emph{otterhound}, they are all dog breeds, and the salient feature sets are focused primarily on the dog. 
This result may suggest that the model has learned generalizable features associated with dogs\,---\,a positive characteristic if we plan to deploy this model.

\input{figures/human_annotation}

Since the model learned to associate the entire dog region with dog classes, this prompts a follow-up question: how much of the dog do we have to annotate before the model no longer associates it with dog breed classes?
Brushing over just the dog's head (Figure~\ref{fig:human-annotation-example}b) or even just the dog's snout (Figure~\ref{fig:human-annotation-example}c) still returns dog breeds as the top classes.
This result suggests the model has learned to correlate even small characteristic features (e.g., black noses) with dogs.

This style of analysis also enables us to ask questions about other objects in the image. 
Although the model was trained to classify this image as \emph{otterhound}, it was also trained to classify 1,000 ImageNet objects.
Thus, the model may know salient information about other objects in the image as well.
In Figure~\ref{fig:human-annotation-example}d we validate this claim by brushing the person's hat and observing the top returned classes are types of hats: \emph{sombrero}, \emph{cowboy hat}, and \emph{bonnet}.
Similarly, we select the person's hand (Figure~\ref{fig:human-annotation-example}e) and, as \emph{hand} is not a class our model was trained to detect, observe classes associated with hands such as \emph{cleaver}, \emph{notebook}, and \emph{space bar}.
This result is intriguing because a hand is often, but not always, present in images of these objects. 
Thus, further analysis is warranted to determine if the model is overly-reliant on the presence of hands to make predictions for these classes.

We can also probe the model to see if it has learned anything about image backgrounds or textures, despite only being trained on foreground objects.
In Figure~\ref{fig:human-annotation-example}f, we select a region of the stone wall.
Interestingly, the model returns classes associated with rocks such as \emph{cliff}, suggesting that training on images with foreground labels may still impart information to the model about background scenes.

As we have seen, Shared Interest allows us to probe model behavior in new ways, enabling exploration into what the model has learned and where it might fail.
Users can identify subsets of features important to classification, explore how well a model can identify secondary objects, and even study the extent to which a model has learned about objects it has never classified.
Using this procedure can help a user test hypotheses about what the model has learned and identify information that could help them improve model behavior.

%% file: figures/cv_prototype.tex
\begin{figure*}[t]
    \begin{center}
    \centerline{\includegraphics[width=\linewidth]{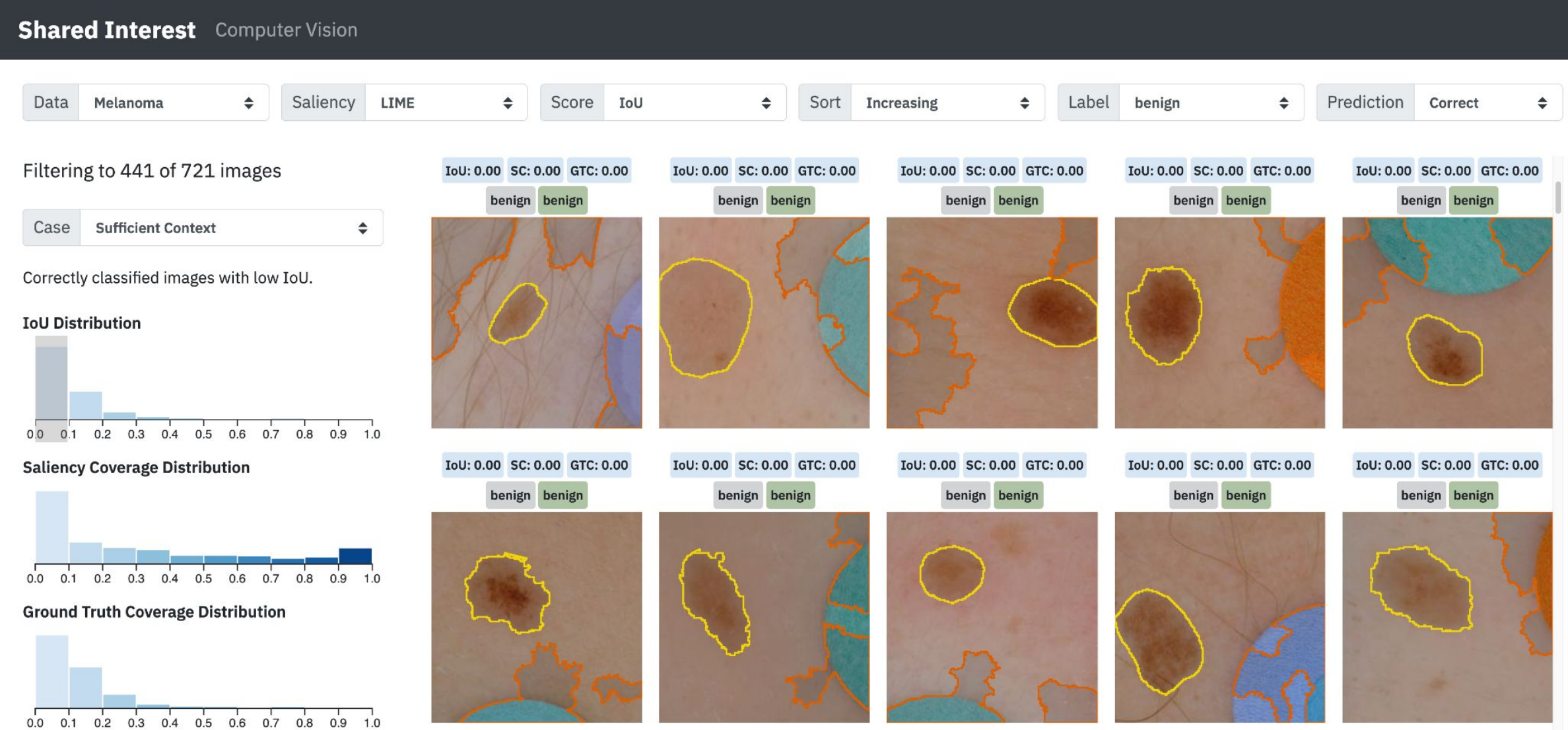}}
    \caption{Shared Interest can help domain experts decide the degree to which they trust a particular model. We use Shared Interest with a dermatologist to analyze a model trained to predict melanoma. The computer vision prototype displays lesion images with segmentations (yellow), \textsc{LIME} explanations (orange), actual (grey) and predicted classifications (green if correct, red otherwise), and all three Shared Interest scores. It enables efficient visual analysis, even by non-expert users, by filtering and sorting based on score, case, label, and prediction. The \textsc{sufficient context} case, shown here, surfaces images where the model has latched onto artifacts to make a \emph{benign} prediction. Since these artifacts only occur in \emph{benign} dataset images, they are sufficient to make a prediction; however, this model would not generalize to clinical cases where the artifacts may also occur in \emph{malignant} images. This demo is at: \url{http://shared-interest.csail.mit.edu/computer-vision/}}
    \label{fig:cv-prototype}
    \Description{The Shared Interest computer vision interface is populated with dermatology images. On the top, there are dropdowns used to select the data (Melanoma), saliency method (LIME), score (IoU), sort (increasing), label (benign), and prediction (correct). The left panel contains a case selection and IoU, Saliency Coverage, and Ground Truth Coverage distributions. The sufficient context case is selected, and 0.0 to 0.1 is highlighted on the IoU distribution. The central panel shows ten images of lesions that fit the filtering criteria. Each lesion shown has 0.0 for all Shared Interest scores and is correctly predicted as benign. Nine images contain purple, green, and blue artifacts beside the lesions. The saliency and ground truth regions for all images are disjoint, with the saliency primarily focused on the artifacts.}
    \end{center}
\end{figure*}

%% file: figures/nlp_prototype.tex
\begin{figure*}[t]
    \begin{center}
    \centerline{\includegraphics[width=\linewidth]{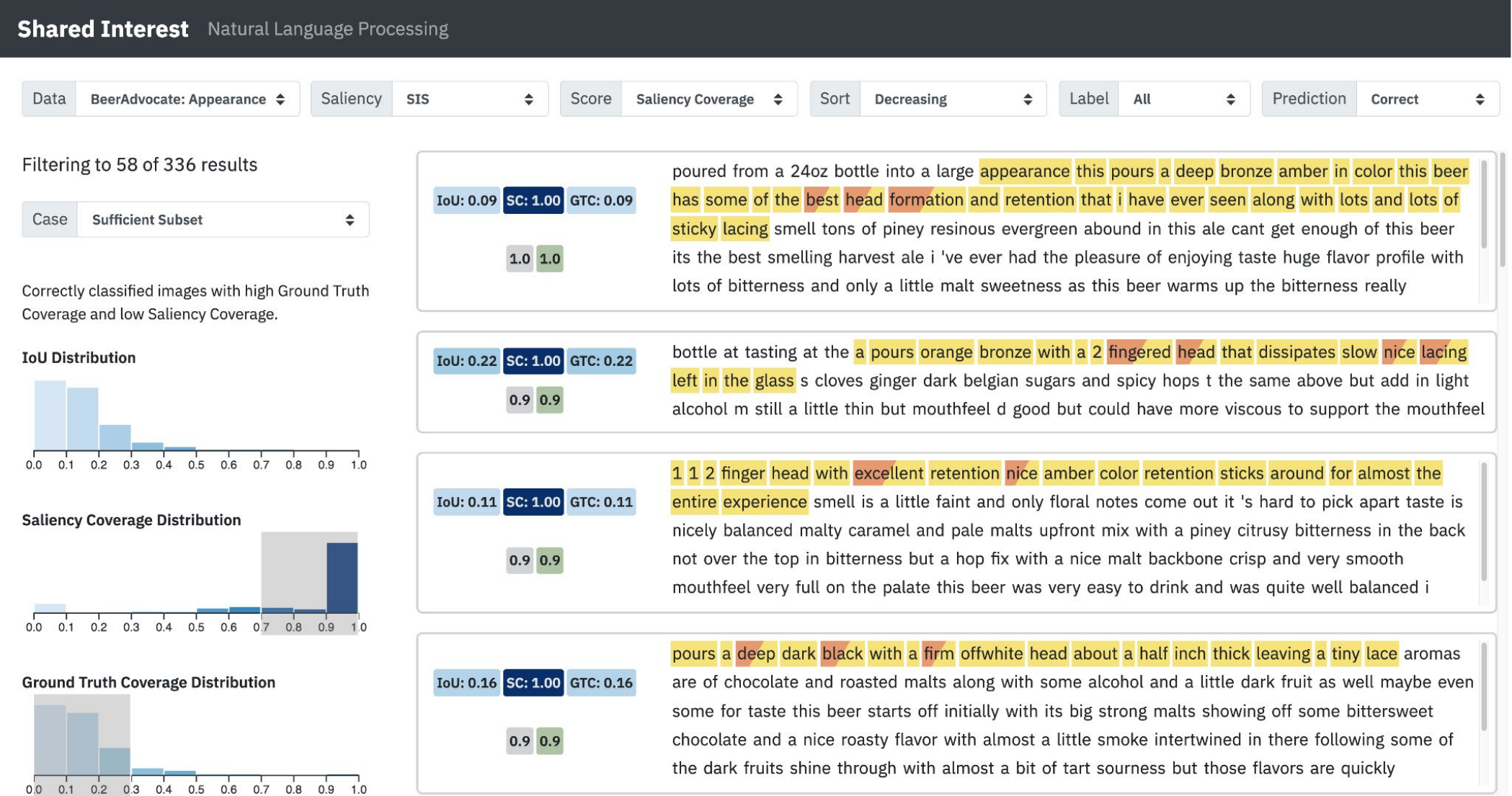}}
    \caption{Shared Interest can assist machine learning experts by enabling efficient large-scale analyses of their methods. We evaluate Shared Interest with a saliency method researcher by assessing the method's performance on a model trained to predict sentiment on text reviews. The NLP prototype displays reviews from the BeerAdvocate dataset with ground truth features (yellow) and SIS saliency features (orange) highlighted. Each review is annotated with its Shared Interest scores, label (grey), and prediction (green if correct, red otherwise). The \textsc{sufficient subset} case, shown here, identifies reviews where the saliency method indicates the model relied on meaningful features, such as ``best head formation'', and where it overfits to general positive sentiment words such as ``excellent'' and ``nice''. This demo is at: \url{http://shared-interest.csail.mit.edu/nlp/}}
    \label{fig:nlp-prototype}
    \Description{The Shared Interest NLP interface populated with review text. On the top there are dropdowns used to select the data (BeerAdvocate: Appearance), saliency method (SIS), score (Saliency Coverage), sort (decreasing), label (all), and prediction (correct). The left panel contains a case selection and IoU, Saliency Coverage, and Ground Truth Coverage distributions. The sufficient subset case is selected, 0.7 to 1.0 is highlighted on the Saliency Coverage distribution, and 0.0 to 0.3 is highlighted on the Ground Truth Coverage distribution. The central panel shows four beer reviews that fit the filtering criteria. Each review shown has 1.0 for its Saliency Coverage score and is correctly predicted. The saliency features are subsets of the ground truth features in all reviews.}
    \end{center}
\end{figure*}

%% file: figures/human_annotation_prototype.tex
\begin{figure}[t]
    \begin{center}
    \centerline{\includegraphics[width=\linewidth]{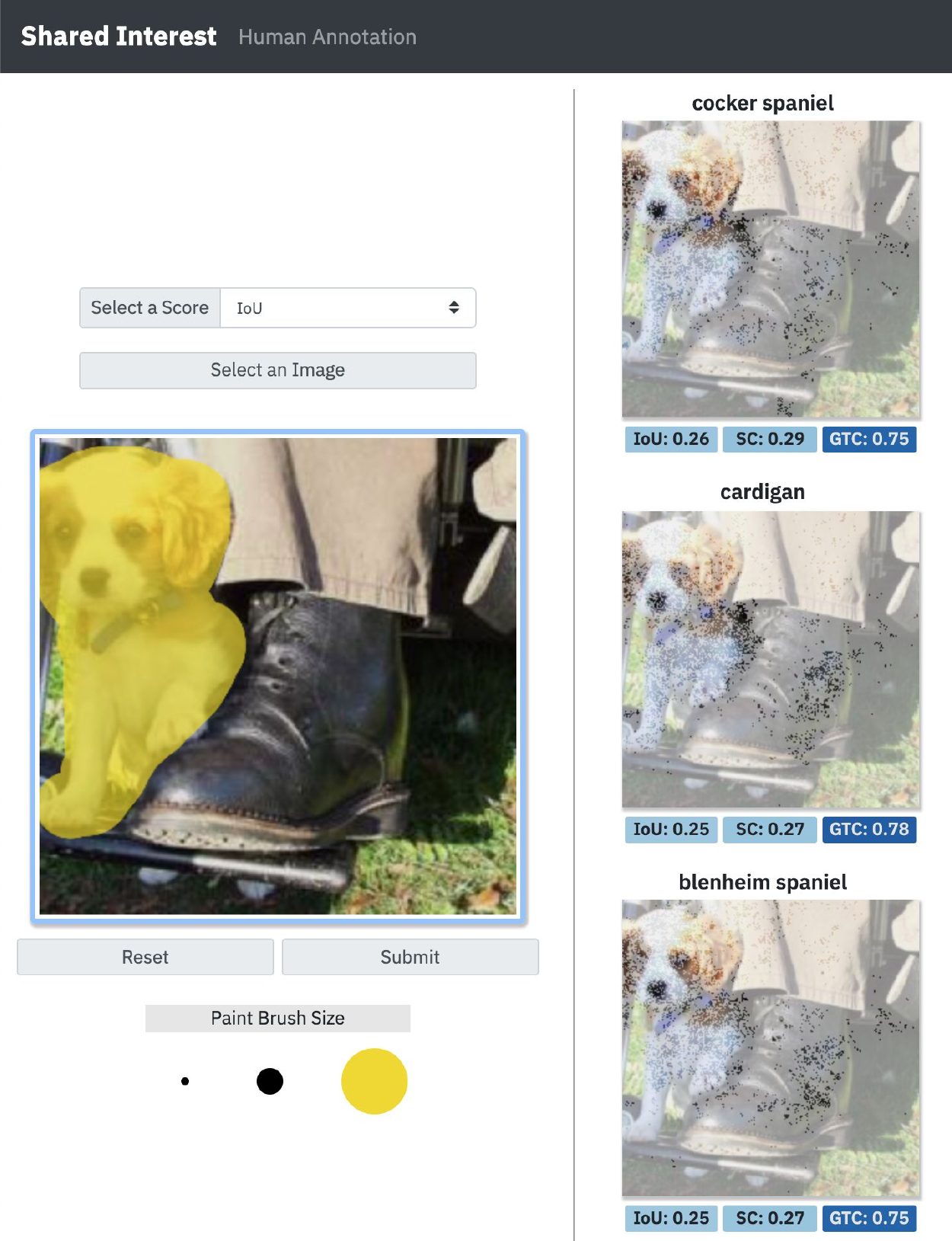}}
    \caption{Shared Interest permits visual workflows where users can interactively probe and explore model behavior. We use the prototype to evaluate an ImageNet classification model. By comparing the user's annotation (shown in yellow) to the saliency feature set (shown via saturation) for every ImageNet class, Shared Interest surfaces the classes most related to the annotated features. In this example, the model relates the dog breed classes \emph{cocker spaniel}, \emph{cardigan}, and \emph{blenheim spaniel} to the features of the highlighted dog. This demo is at: \url{http://shared-interest.csail.mit.edu/human-annotation/}}
    \label{fig:human-annotation-prototype}
    \Description{The Shared Interest human annotation interface contains a selection pane and a results pane. The selection pane shows an image of a dog beside a person's shoe, annotation tools, and a score selection. The dog has been annotated, and IoU is the selected score. The results pane shows the same image with saliency features overlaid for three classes. The top classes are Cocker Spaniel, Cardigan, and Blenheim Spaniel. The saliency features for each class focus mainly on the dog.}
    \end{center}
\end{figure}

%% file: figures/human_annotation.tex
\begin{figure*}[t]
    \begin{center}
    \centerline{\includegraphics[width=\linewidth]{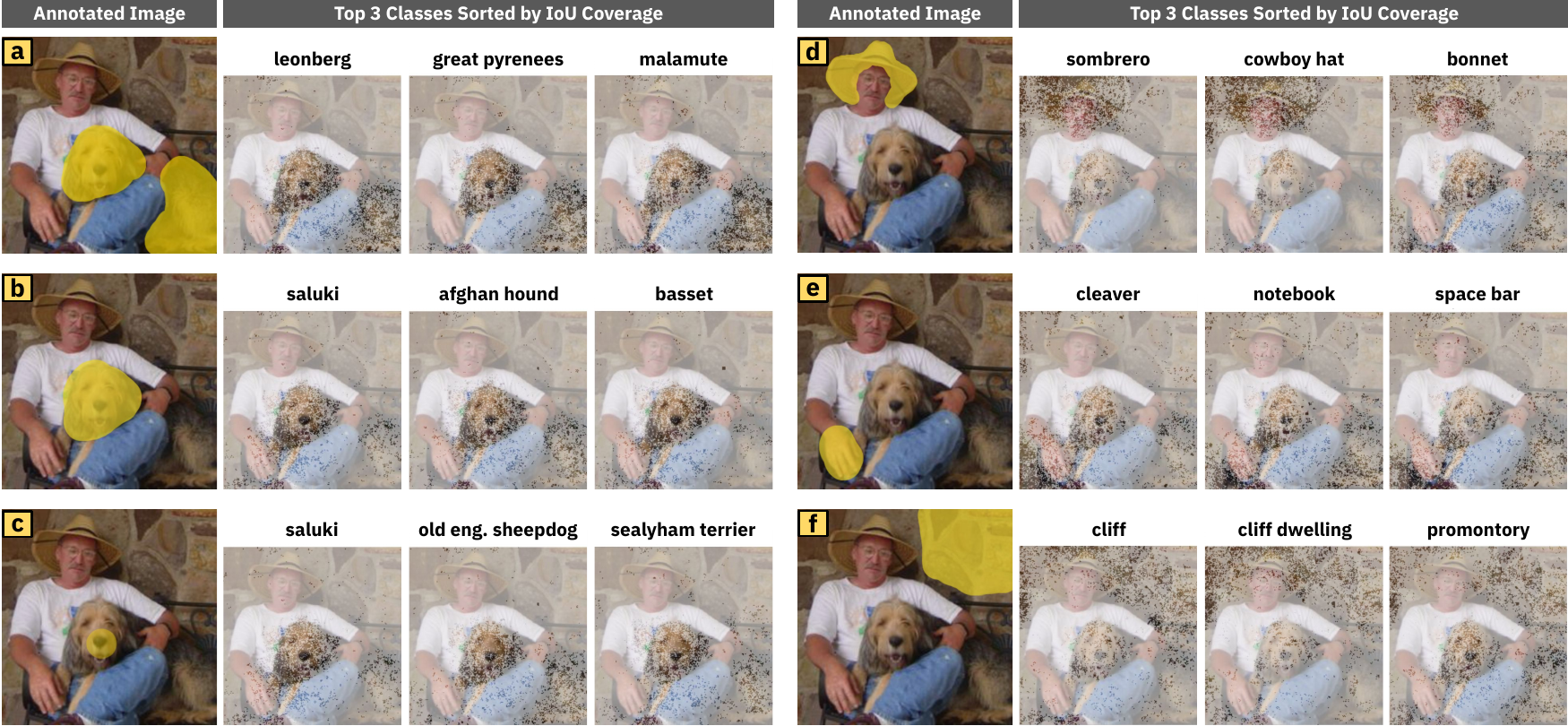}}
    \caption{Shared Interest enables users to probe the model with different ground truth features to understand model behavior. By comparing the user's annotation (shown in yellow) to the saliency feature set (shown via saturation) for every ImageNet class, Shared Interest surfaces the classes most related to the annotated features. Probing with smaller and smaller sets of features (a-c) shows the model has learned characteristic features of dogs. Probing with secondary objects (d) demonstrates that the model learns about objects other than the labeled object in the image. Probing the model with features it has not learned to classify (e) indicates the model learns to relate these features to associated objects (e.g., hand and cleaver). Finally, probing with background features (f) demonstrates that the model has learned related features despite only being trained on foreground classification.}
    \label{fig:human-annotation-example}
    \Description{Six examples, labeled a-f, show an annotated image paired with results of the top three classes sorted by IoU coverage. The image is of a man and a dog sitting on a bench in front of a rock wall. The results show the same image annotated with the saliency and the class name. (a) shows the dog's head and body annotated. The results are Leonberg, Great Pyrenees, and Malamute. The saliency features highlight the dog's head and body. (b) shows the dog's head annotated. The results are Saluki, Afghan Hound, and Basset. The saliency features focus on the dog's face. (c) shows the dog's nose annotated. The results are Saluki, Old English Sheepdog, and Sealyham Terrier. The saliency features are concentrated on the dog's nose. (d) shows the man's hat annotated. The results are sombrero, cowboy hat, and bonnet. The saliency features are concentrated on the hat. (e) shows the man's hand annotated. The results are cleaver, notebook, and space bar. The saliency features focus on the man's hand with some features elsewhere in the image. (f) shows the rock wall annotated. The results are cliff, cliff dwelling, and promontory. The saliency features are concentrated on the rock wall with some features scattered elsewhere in the image.}
    \end{center}
\end{figure*}

%% file: sections/06_discussion.tex
\section{Discussion}
This paper presents Shared Interest, a method for large-scale analysis of machine learning model behavior via metrics that quantify instances based on the model's alignment with human reasoning.
Shared Interest enables instances to be sorted, ranked, and aggregated based on this alignment. 
Using Shared Interest, we identified eight patterns in model behavior that recur across multiple domains (computer vision and natural language processing), model architectures (convolutional and recurrent neural networks), and saliency methods (gradient-based and model-agnostic).
These patterns range from cases where the ground truth features are important to the model's incorrect prediction (\textsc{confuser}) to cases where the ground truth features are not important to the model's correct classification (\textsc{sufficient context}).
We evaluate Shared Interest's usefulness through representative case studies of real-world interactive visual analysis workflows.
Working with a dermatologist and a machine learning researcher revealed that although analysts want to explore model behavior, they find current methods are tedious to use and require too much ad hoc inspection to feel entirely confident in the results. 
In contrast, with Shared Interest, both types of users could systematically explore model behavior to identify reasons to question the model's reliability and validate novel saliency methods.
In a final case study, we demonstrate that Shared Interest is not restricted to merely understanding a model's predictive performance but can also support interactive ``what if'' analysis to determine the input features most important to particular predictions.

\subsection{Limitations}
While the Shared Interest methodology can help users efficiently and comprehensively understand model behavior, it requires data paired with ground truth annotations.
Research datasets, such as those used in this paper may include such annotations, but real-world data rarely do due to the time and effort required in the collection process.
While this issue can limit Shared Interest's applicability, we believe that understanding model behavior is critical enough to warrant the collection of human annotations.
Collection may range from annotating a few instances (e.g., via the probing interface) for general research analysis to annotating entire datasets when deciding to deploy a model on a critical task.

Additionally, existing ground truth annotations often highlight the features associated with the label, such as the pixels corresponding to the dog in an image.
However, human decision-making may not perfectly align with those features.
A human may only need to look at a subset of the features like the dog's face to know the image contains a dog.
Alternatively, a human may need additional features like a hockey player or ice rink to know that a black round object is a hockey puck.
Thus, as more work focuses on understanding human decision-making and annotating datasets in a corresponding rich fashion, Shared Interest metrics will more precisely communicate human-AI alignment.

Finally, throughout this paper, we rely on saliency methods as proxies for model reasoning.
However, researchers have demonstrated cases where saliency methods do not accurately reflect the model~\citep{adebayo2018sanity, kindermans2019reliability}.
While saliency methods are valuable tools that can give insight into model behavior, Shared Interest can inherit their limitations. 
For instance, if a saliency method returns an explanation that does not accurately reflect the model's decision-making, Shared Interest will not be able to quantify human-AI alignment accurately.
Nonetheless, we designed Shared Interest to be agnostic to the saliency method; so, as methods evolve, Shared Interest's ability to communicate model-human alignment will also improve.

\subsection{Future Work}
Shared Interest opens the door to several promising directions for future work.
One straightforward path is applying Shared Interest to tabular data\,---\,a standard format used to train models, particularly in healthcare applications.
Tabular data is often more semantically complex than image or text data and thus allows us to bring further nuance to the recurring behavior patterns we have identified in this paper. 
For instance, fields in tabular data may correlate in more specific and fine-grained ways (e.g., as proxy variables~{\citep{mehrabi2019survey}}) than the foreground/background context we have distinguished in this paper. 
As tabular data for these uses cases often contains sensitive personal information (e.g., health data), one could imagine using a version of our interactive probing prototype to systematically analyze how a particular model may perpetuate or amplify bias in the data.

Another avenue for future work is using Shared Interest to compare the fidelity of different saliency methods.
Previous work has conducted experiments to determine how faithful saliency methods are to an underlying model~{\citep{adebayo2018sanity}} based on cascading randomization of internal model layers and its effect, or lack thereof, on the resulting saliency map.
These prior studies compute quantitative metrics to identify divergences between perturbations to the model and the effect on the saliency.
However, these metrics operate only over individual pixels. 
By rerunning these studies and quantifying results in terms of Shared Interest metrics, we can increase the level of abstraction of the results.
For instance, rather than defining input invariance~{\citep{kindermans2019reliability}} over individual pixels, we could define it over GTC, SC, or IoU and distinguish whether saliency map sensitivity represents a semantically meaningful signal.

Finally, an exciting direction might consider Shared Interest during model training.
Currently, model developers have limited introspection into this process.
Typically, they rely on curves that visualize a model's loss function or performance on the training and validation sets. 
While visual analytics research has helped, most existing work has focused on depicting model architecture or performance~\citep{kahng2017cti, ren2016squares, wongsuphasawat2017visualizing}.
Shared Interest, however, allows us to evaluate training in terms of the model's reasoning.
By comparing how saliency features change over epochs, Shared Interest could identify the instances a model gets right immediately and the instances that take several more updates to classify correctly. 
These insights could inform future training procedures or augment the dataset with more informative examples.

%% file: sections/07_appendix.tex
\renewcommand{\thefigure}{\Alph{figure}}
\setcounter{figure}{0}

\renewcommand{\thetable}{\Alph{table}}
\setcounter{table}{0}
\clearpage

\section{Additional Saliency Methods}
\label{sec:appendix-saliency}
The Shared Interest metrics represent model decision-making using a saliency feature set computed via a saliency method.
Throughout the paper, we show that Shared Interest can be used with a variety of saliency methods, including Vanilla Gradients~\citep{simonyan2013deep}, Integrated Gradients~\citep{sundararajan2017axiomatic}, LIME~\citep{ribeiro2016should}, and SIS~\citep{carter2019made}.

Here we explore additional saliency methods: SmoothGrad~\citep{smilkov2017smoothgrad}, Guided Backpropagation~\citep{springenberg2014striving}, Gradient SHAP~\citep{lundberg2017unified}, and Grad-Cam~\citep{selvaraju2017grad}.
For each saliency method we show examples of the Shared Interest cases (Figure~\ref{fig:cases-per-saliency}) and the distribution of Shared Interest scores (Figure~\ref{fig:thresholding-distributions}) across vehicle images from ImageNet-9~\citep{tsipras2020imagenet}.
Implementations used to compute the saliency methods are available at \url{https://github.com/mitvis/shared-interest}.
The SmoothGrad implementation is based on the Google PAIR implementation: \url{https://github.com/PAIR-code/saliency}, and all other methods are implemented using Captum~\citep{kokhlikyan2020captum}.
In each example, we compute attribution with respect to the predicted class.

We find that the Shared Interest cases occur regardless of saliency method, indicating Shared Interest can be successfully used with various saliency methods.
While each saliency method highlights the features important to the model's decision, each method computes importance differently, and the outputted saliency features vary across saliency methods.
For example, Vanilla Gradients represents importance as the impact a slight change in each feature would have on the model's output, and it often results in sparse and noisy feature subsets.
On the other hand, GradCAM computes the gradients with respect to the last convolutional layer, which results in continuous feature regions.
The Shared Interest scores are often higher for continuous feature regions than sparse feature sets (see Figure~\ref{fig:thresholding-distributions}).
Thus, Shared Interest scores should only be compared within a single method because each method's score distribution will vary slightly due to variations in the methods.
Further, the high and low values selected for each Shared Interest case will also vary based on the saliency method.

\input{figures/cases_per_saliency_method}

\section{Additional Discretization Techniques}
\label{sec:appendix-thresholding}

To utilize Shared Interest's set-based metrics, the saliencies from methods that output continuous-valued scores must be discretized.
In this paper, we show that Shared Interest surfaces insights into model behavior using a variety of discretization techniques, including score-based thresholding and model-based thresholding (see Section~\ref{sec:experiment-details}).
In Figure~\ref{fig:thresholding-example} we show additional score-based discretization techniques, and in Figure~\ref{fig:thresholding-distributions}, we show the Shared Interest score distributions for each technique.
These techniques threshold values based on the saliency scores.
Some methods take features whose scores are above a particular value (i.e., mean, one standard deviation above the mean, and two standard deviations above the mean).
Other methods take the top $n$ features, such as the top 5\% - 75\% of features or the same number of saliency features as ground truth features.

\input{figures/thresholding_examples}

Many discretization techniques can successfully be used in Shared Interest, but each technique makes its own assumptions about the model and saliency method.
The Shared Interest metrics depend on the thresholding technique, and the distribution of Shared Interest scores will vary based on the threshold.
For example, assuming ground truth features are also the most salient features, stricter thresholds that result in fewer saliency features will increase SC and decrease GTC and IoU.
Thus, Shared Interest scores should only be compared within a single discretization technique.
The ``high'' and ``low'' Shared Interest values used to compute the Shared Interest cases also depend on the discretization procedure and should be chosen based on the score distribution.
Finally, size-based thresholds (e.g., features with the top 25\% of saliency values) artificially determine the number of saliency features and cause Shared Interest metrics to vary across instances depending on the number of ground truth features.
For example, if the method results in a small saliency feature set, instances with large ground truth feature sets will have low IoU and GTC scores.
The metrics will be comparable as long as the ground truth feature sets are similar in size across instances (e.g., normalized medical images).
However, if the number of ground truth features varies significantly across the dataset, a thresholding technique based on the saliency value or the model's behavior is preferred.

\input{figures/thresholding_distributions}

%% file: figures/cases_per_saliency_method.tex
\begin{figure*}[h]
    \begin{center}
    \centerline{\includegraphics[width=\linewidth]{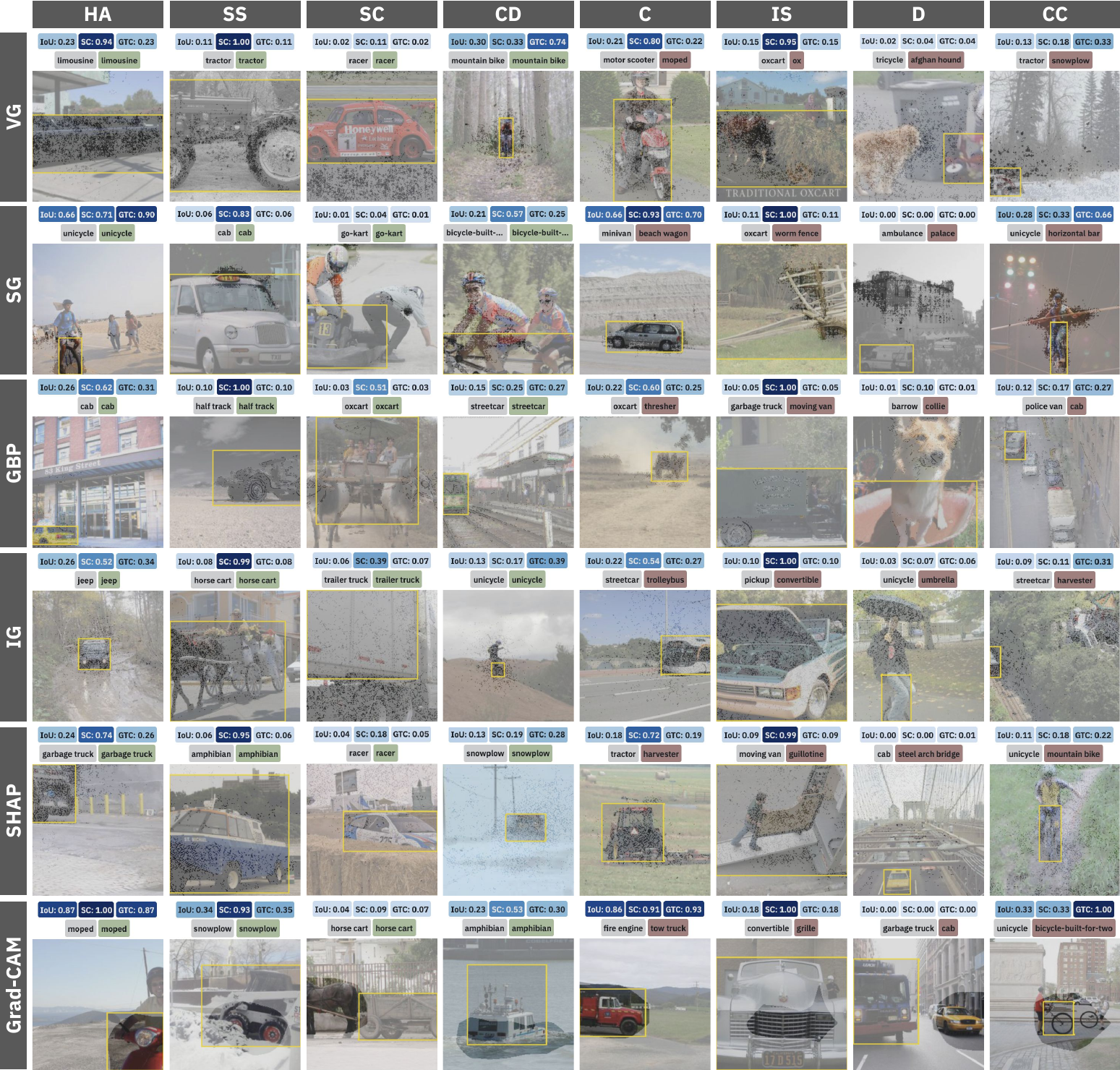}}
    \caption{Shared Interest is agnostic to saliency method, and the Shared Interest cases occur across a variety of saliency methods. Here we show an example of each of the eight Shared Interest cases\,---\,\textsc{human aligned} (HA), \textsc{sufficient subset} (SS), \textsc{sufficient context} (SC), \textsc{context dependent} (CD), \textsc{confuser} (C), \textsc{insufficient subset} (IS), \textsc{distractor} (D), and \textsc{context confusion} (CC)\,---\,across saliency methods\,---\,Vanilla Gradients (VG)~\citep{simonyan2013deep}, SmoothGrad (SG)~\citep{smilkov2017smoothgrad}, Guided Backpropagation (GBP)~\citep{springenberg2014striving}, Integrated Gradients (IG)~\citep{sundararajan2017axiomatic}, Gradient SHAP (SHAP)~\citep{lundberg2017unified}, and Grad-CAM~\citep{selvaraju2017grad}. Images are vehicle images in ImageNet-9~\citep{deng2009imagenet} and the saliency methods are thresholded at one standard deviation above the mean. Each image is annotated with the label (grey), prediction (green if correct, red otherwise), and Shared Interest scores. The ground truth features are shown in yellow and the saliency features are shown via saturation.}
    \label{fig:cases-per-saliency}
    \Description{A grid of example images from each Shared Interest case for different saliency methods. The columns are Shared Interest cases: human aligned, sufficient subset, sufficient context, context dependent, confuser, insufficient subset, distractor, and context confusion. The rows are saliency methods: Vanilla Gradients, SmoothGrad, Guided Backpropagation, Integrated Gradients, SHAP, and Grad-CAM. Each item shows an image with its saliency and ground truth feature sets, label and prediction, and Shared Interest scores.}
    \end{center}
\end{figure*}

%% file: figures/thresholding_examples.tex
\begin{figure*}[h]
    \begin{center}
    \centerline{\includegraphics[width=\linewidth]{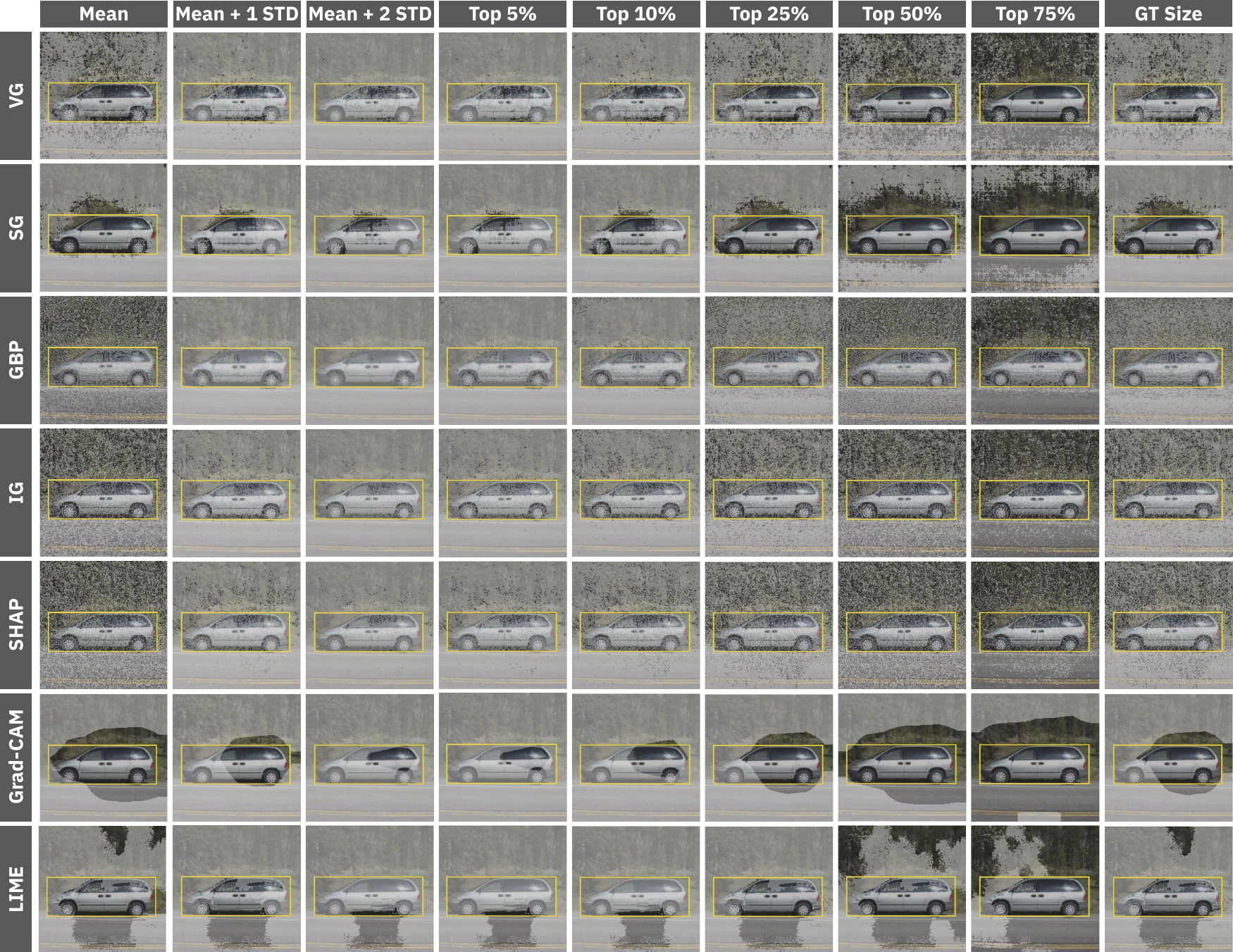}}
    \caption{Many discretization techniques can be successfully used with Shared Interest. Discretization can be score-based (dependent on the saliency scores) or model-based (dependent on the model's output). Here we show score-based thresholding techniques across different saliency methods on an ImageNet~\citep{deng2009imagenet} \textit{minivan} image. We compare Vanilla Gradients (VG)~\citep{simonyan2013deep}, SmoothGrad (SG)~\citep{smilkov2017smoothgrad}, Guided Backpropagation (GBP)~\citep{springenberg2014striving}, Integrated Gradients (IG)~\citep{sundararajan2017axiomatic}, Gradient SHAP (SHAP)~\citep{lundberg2017unified}, Grad-CAM~\citep{selvaraju2017grad}, and LIME~\citep{ribeiro2016should}. Score-based thresholding can be performed on the saliency values (e.g., all features with a saliency value greater than the mean saliency value) or based on the number of features (e.g., all features whose saliency is in the top 10\% of saliency values). In this example, we compare thresholds of the mean, one standard deviation above the mean, two standard deviations above the mean, top 5\%-75\% of saliency values, and the same number of features as ground truth features (GT Size). As the thresholding technique relaxes, more features are added to the saliency feature set (shown via saturation), which changes the relationship to the ground truth features (shown in yellow). The Shared Interest scores depend on the discretization, and analysis should be performed with the discretization procedure's assumptions in mind.}
    \label{fig:thresholding-example}
    \Description{A grid of images comparing saliency outputs at different discretization techniques. The columns are thresholding techniques: mean, mean plus one standard deviation, mean plus two standard deviations, top 5\% of values, top 10\% of values, top 25\% of values, top 50\% of values, top 75\% of values, and the same number of features as ground truth features. The rows are saliency methods: Vanilla Gradients, SmoothGrad, Guided Backpropagation, Integrated Gradients, Gradient SHAP, Grad-CAM, and LIME. Each item shows an ImageNet minivan image with a yellow ground truth bounding box and the saliency overlaid using saturation. As the thresholding technique increases the threshold or number of features, more of the minivan is indicated as important.}
    \end{center}
\end{figure*}

%% file: figures/thresholding_distributions.tex
\begin{figure*}[h]
    \begin{center}
    \centerline{\includegraphics[width=\linewidth]{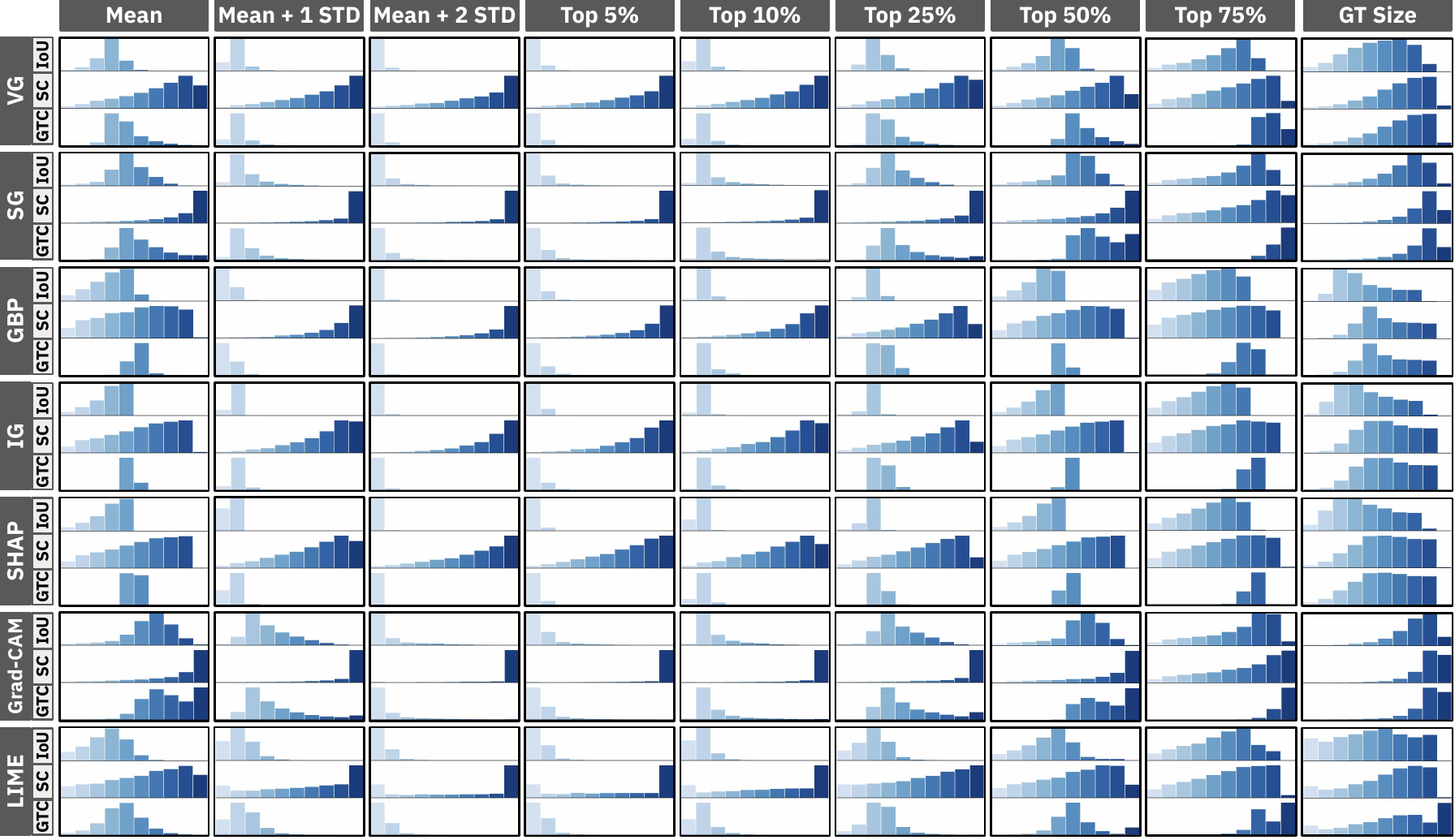}}
    \caption{Discretization techniques directly impact the Shared Interest scores. Here we compare the IoU, SC, and GTC Shared Interest score distributions on vehicle images from the ImageNet-9 dataset~\citep{tsipras2020imagenet}. We compare distributions across saliency methods --- Vanilla Gradients (VG)~\citep{simonyan2013deep}, SmoothGrad (SG)~\citep{smilkov2017smoothgrad}, Guided Backpropagation (GBP)~\citep{springenberg2014striving}, Integrated Gradients (IG)~\citep{sundararajan2017axiomatic}, Gradient SHAP (SHAP)~\citep{lundberg2017unified}, Grad-CAM~\citep{selvaraju2017grad}, and LIME~\citep{ribeiro2016should} --- and thresholding technique --- mean, one standard deviation above the mean, two standard deviations above the mean, top 5\%-75\% of saliency values, and the same number of features as ground truth features (GT Size). As the thresholding technique becomes stricter, IoU and GTC decrease and SC increases.}
    \label{fig:thresholding-distributions}
    \Description{A grid of Shared Interest score distributions comparing saliency outputs at different discretization techniques. The columns are thresholding techniques: mean, mean plus one standard deviation, mean plus two standard deviations, top 5\% of values, top 10\% of values, top 25\% of values, top 50\% of values, top 75\% of values, and the same number of features as ground truth features. The rows are saliency methods: Vanilla Gradients, SmoothGrad, Guided Backpropagation, Integrated Gradients, Gradient SHAP, Grad-CAM, and LIME. Each item shows the IoU Coverage, Saliency Coverage, and Ground Truth Coverage distributions. As the thresholding technique increases, the threshold or number of features, IoU Coverage and Ground Truth Coverage decrease, and Saliency Coverage increases.}
    \end{center}
\end{figure*}